\documentclass[submission,copyright,creativecommons]{eptcs}
\providecommand{\event}{QPL 2020} 
\usepackage{breakurl}             
\usepackage{underscore}
\usepackage[english]{babel}
\usepackage[utf8x]{inputenc}
\usepackage[T1]{fontenc}
\usepackage{cite}
\usepackage{xcolor}
\usepackage{bbold}
\usepackage{subfig}
\usepackage{enumitem}
\usepackage{comment}
\usepackage{amssymb}
\usepackage{amsfonts}

\usepackage{amsmath}
\usepackage{amssymb}
\usepackage{graphicx}
\usepackage{physics}
\usepackage{tensor}
\usepackage{mathtools}
\usepackage{proof} 
\usepackage{prftree}
\usepackage{booktabs}
\usepackage{stmaryrd}
\usepackage{todonotes}
\usepackage{comment}
\usepackage{amsthm}
\usepackage{ragged2e}

\theoremstyle{definition}
\newtheorem{definition}{Definition}[section]

\newcommand{\cut}[1]{} 
\newcommand{\bs}{\backslash}
\newcommand{\fdia}{\lozenge}
\newcommand{\gbox}{\square}
\newcommand{\cdott}{\>\cdot\>}
\newcommand{\bo}{[}
\newcommand{\bc}{]}

\newcommand{\Dg}[1]{\llbracket #1 \rrbracket_g}
\newcommand{\Spc}[1]{\lceil #1 \rceil}
\newcommand{\Hm}[1]{\lfloor #1 \rfloor}
\newcommand{\nd}[2]{#1\vdash #2}

\title{Putting a Spin on Language:\\ A Quantum Interpretation of Unary Connectives for Linguistic Applications}
\author{Adriana D. Correia \qquad\qquad Henk T. C. Stoof
\institute{Institute for Theoretical Physics\\ Center for Complex Systems Studies}
\institute{Utrecht University\\
Utrecht, The Netherlands}
\email{\quad a.duartecorreia@uu.nl \quad\qquad h.t.c.stoof@uu.nl}
\and
Michael Moortgat
\institute{Utrecht Institute of Linguistics OTS\\ Center for Complex Systems Studies}
\institute{Utrecht University\\
Utrecht, The Netherlands}
\email{m.j.moortgat@uu.nl}}
\def\titlerunning{Putting a Spin on Language}
\def\authorrunning{A. D. Correia, H. T. C. Stoof and M. Moortgat}
\begin{document}
\maketitle

\begin{abstract}
Extended versions of the Lambek Calculus currently used in computational linguistics rely on
unary modalities to allow for the controlled application of structural rules affecting
word order and phrase structure. These controlled structural operations give rise to
derivational ambiguities that are missed by the original Lambek Calculus or its pregroup simplification.
Proposals for compositional interpretation of extended Lambek Calculus in the compact closed
category of FVect and linear maps have been made, but in these proposals the syntax-semantics
mapping ignores the control modalities, effectively restricting their role to the syntax.
Our aim is to turn the modalities into first-class citizens of the vectorial interpretation.
Building on the directional density matrix semantics, we extend the interpretation
of the type system with an extra spin density matrix space. The interpretation of proofs then
results in ambiguous derivations being tensored with orthogonal spin states. Our method introduces
a way of simultaneously representing co-existing interpretations of ambiguous utterances, and
provides a uniform framework for the integration of lexical and derivational ambiguity.
\end{abstract}

\section{Introduction}

A cornerstone of formal semantics is Montague's \cite{MontagueUG} compositionality theory.
Compositional interpretation, in this view, is a homomorphism, a structure-preserving map
that sends types and derivations of a syntactic source logic to the corresponding
semantic spaces and operations thereon. In the DisCoCat framework \cite{coecke2010mathematical}
compositionality takes a surprising new turn. Montague's abstract mathematical view on the
syntax-semantics interface is kept, but the non-committed view on \emph{lexical}
meaning that one finds in formal semantics is replaced by a data-driven, distributional
modelling, with finite dimensional vector spaces and linear maps as the target
for the interpretation function. More recently density matrices and completely positive maps have been used to treat lexical ambiguity \cite{piedeleuopen}, word and sentence entailment \cite{DBLP:journals/amai/SadrzadehKB18, bankova2019graded} and meaning updating \cite{coecke2020meaning}.

Our goal in this paper is to apply the DisCoCat methodology to an extended version
of the Lambek calculus where structural rules affecting word order and/or phrase
structure are no longer freely available, but have to be explicitly licensed by
unary control modalities, \cite{mmjolli,KurtoMM}. In particular, we adjust the interpretation homomorphism
to assign appropriate semantic spaces to the modally extended type language,
and show what their effect is on the derivational semantics. We choose to use density matrices as our interpretation spaces and show that, besides allowing for an integration of our model with other forms of ambiguity at the lexical level, it is key to preserve information about the ambiguity at phrase level. 

The paper is structured as follows. In section \ref{types} we recall the natural deduction
rules of the simply typed Lambek Calculus, with the associated lambda terms under the
proofs-as-programs interpretation. We extend the language with a residuated pair of
unary modalities $\fdia, \gbox$ and show how these can be used to control structural reasoning,
in particular reordering (commutativity). As an illustration, we show how the extended type
logic allows us to capture derivational ambiguities that arise in Dutch relative clause constructions.
In section \ref{interpretationspaces} we set up the mapping from syntactic types to semantic spaces,
adding an extra spin space to the previously used density matrix spaces. We motivate the introduction
of this extra space and relate the interpretation of the connectives in these spaces to the measurement
and evolution postulates of quantum mechanics.
In section \ref{operationalinterpretation} we show how the interpretation of the logical
and structural inference rules of our extended type logic accommodates the spin space. In section \ref{spinspace} we make explicit the two-level spin space that we will use to store the ambiguity in the case of Dutch relative clauses.
In section \ref{dutchrelclauses} we return to our example of derivational ambiguity and
show how orthogonal spin states keep track of co-existing interpretations.

\section{Extended Lambek Calculus}\label{types}

By \textbf{NL}$_\fdia$ we designate the (non-associative, non-commutative, non-unital) pure
residuation logic of \cite{Lam61}, extended with a pair of unary type-forming operators
$\fdia,\gbox$, also forming a residuated pair. Formulas are built over a set of
atomic types $\mathcal{A}$ (here \emph{s, np, n} for sentences, noun phrases and common nouns respectively)
by means of a binary product $\bullet$ with its left and right residuals $\slash$, $\backslash$, 
and a unary $\lozenge$ with its residual $\square$:

$$\mathcal{F} ::= \mathcal{A} \mid \square \mathcal{F} \mid \lozenge \mathcal{F} \mid \mathcal{F} \backslash \mathcal{F} \mid 
\mathcal{F} \slash \mathcal{F} \mid \mathcal{F} \bullet \mathcal{F}.$$ 

Figure \ref{nlprograms} gives the (sequent-style) natural deduction presentation, together with the Curry-Howard term
labelling\footnote{We restrict to the simply typed fragment, ignoring the $\bullet$ operation.}.
Judgements are of the form $\Gamma\vdash B$, with $B$ a formula and $\Gamma$ a structure term with formulas at the leaves.
Antecedent structures are built according to the grammar $\mathcal{S} ::= \mathcal{F} \mid (\mathcal{S}\cdot \mathcal{S}) \mid \langle \mathcal{S} \rangle$.
The binary structure-building operation $(-\cdot-)$ is the structural counterpart of the connective $\bullet$ in the formula language.
The unary structure-building operation $\langle -\rangle$ similarly is the counterpart of $\fdia$ in the formula language.

With term labelling added, an antecedent term $\Gamma$ with leaves $x_1:A_1,\ldots,x_n:A_n$ becomes a typing environment giving
type declarations for the variables $x_i$. These variables constitute the parameters for the program $t$ associated
with the proof of the succedent type $B$.
Intuitively, one can see a term-labeled proof as an algorithm to compute a meaning $t$ of type $B$ with parameters $x_i$ of type $A_i$.
In parsing a particular phrase, one substitutes the meaning of the constants (i.e.~words) that make up for the parameters of this
algorithm.

Notice that the term language respects the distinction between $\slash$ and $\backslash$: we use the `directional'
lambda terms of \cite{wansing1992} with left versus right abstraction and application. The inference rules for 
$\gbox$ and $\fdia$ are reflected in the term language by
${}^\vee,{}^\cup$ (Elimination) and ${}^\wedge,{}^\cap$
(Introduction) respectively. 

\begin{figure}[h]
\begin{center}

\[\textrm{Terms:}\qquad
t,u \mathbin{::=} x \mid \lambda^{r} x.t \mid \lambda^{l} x.t \mid t\triangleleft u \mid u\triangleright t \mid ^\cup t \mid ^\cap t \mid ^\vee t \mid ^\wedge t \mid {}^{c}t 
\]

Typing rules:
\[\infer[Ax]{\nd{x:A}{x:A}}{}\]
\[\infer[I/]{\nd{\Gamma}{\lambda^{r} x.t:B/A}}{\nd{\Gamma\cdott x:A}{t:B}} 
\qquad
\infer[I\bs]{\nd{\Gamma}{\lambda^{l} x.t:A\bs B}}{\nd{x:A\cdott\Gamma}{t:B}}\]
\[\infer[E/]{\nd{\Gamma\cdott\Delta}{t\triangleleft u:B}}{\nd{\Gamma}{t:B/A} & \nd{\Delta}{u:A}}
\qquad
\infer[E\bs]{\nd{\Gamma\cdott\Delta}{u\triangleright t:B}}{\nd{\Gamma}{u:A} & \nd{\Delta}{t:A\bs B}}\]
\[\infer[I\>\square]{\nd{\Gamma}{^{\wedge}t:\square B}}{\nd{\langle \Gamma \rangle}{t: B}}
\qquad
\infer[I\>\lozenge]{\nd{\langle \Gamma \rangle}{^{\cap} t:\lozenge B}}{\nd{\Gamma}{t: B}}\]
\[\infer[E\>\square]{\nd{\langle \Gamma \rangle}{^{\vee}t:B}}{\nd{\Gamma}{t:\square B}}
\qquad
\infer[E\>\lozenge]{\nd{\Gamma[\Delta]}{u[^\cup t/ x]:B}}{\nd{\Delta}{t:\lozenge A} & \nd{\Gamma[\langle x:  A \rangle] }{u:B }}\]
\caption{\small \textbf{NL}$\lozenge$. Proofs and terms. Antecedent structure terms must be non-empty.
Notation $\Gamma[\Delta]$ for structure term $\Gamma$ with substructure $\Delta$.}
\label{nlprograms}
\end{center}
\end{figure}

In addition to the logical rules for $\fdia$ and $\gbox$, we are interested in formulating options for structural reasoning keyed to their presence. 
Consider the postulates expressed by the categorical morphisms of
(\ref{strpost}), or the corresponding inference rules of (\ref{strrules}) in the N.D.~format of Figure \ref{nlprograms}. These represent controlled forms of associativity and commutativity, explicitly licensed by the presence of $\fdia$ (or its structural counterpart $\langle-\rangle$ in the sequent rules).

\begin{equation}\label{strpost}
\fdia A\otimes(B\otimes C)\longrightarrow (\fdia A\otimes B)\otimes C
\qquad
\fdia A\otimes(B\otimes C)\longrightarrow B\otimes(\fdia A\otimes C)
\end{equation}
\begin{equation}\label{strrules}
\infer[Ass_\lozenge]{\nd{\Gamma[\langle \Delta_1 \rangle\cdott(\Delta_2\cdott \Delta_3)]}{t:B}}{\nd{\Gamma[(\langle\Delta_1\rangle\cdott\Delta_2)\cdott \Delta_3]}{t:B} }
\qquad
\infer[Comm_\lozenge]{\nd{\Gamma[\langle \Delta_1 \rangle\cdott(\Delta_2\cdott \Delta_3)]}{
{}^ct:B}}{\nd{\Gamma[\Delta_2\cdott(\langle \Delta_1 \rangle\cdott \Delta_3)]}{t:B} }
\end{equation}

Controlled forms of structural reasoning of this type have been used to model the dependencies between question words
or relative pronouns and `gaps' (physically unrealized hypothetical resources) that follow them. We illustrate
with Dutch relative clauses, and refer the reader to \cite{moortgat2017lexical} for a vector-based semantic analysis. Dutch, like
Japanese, has verb-final word order in embedded clauses as show in (\ref{dutch}a) which translates as (\ref{dutch}b).
Now consider the relative clause (\ref{dutch}c). It has two possible interpretations, expressed by the translations
(\ref{dutch}d) and (\ref{dutch}e). With a typing $(n\bs n)/(np\bs s)$ for the relative pronoun `die' we
can capture only the (\ref{dutch}d) interpretation; the improved typing $(n\bs n)/(\fdia\gbox np\bs s)$
creates a derivational ambiguity that covers both the (\ref{dutch}d) and the (\ref{dutch}e) interpretation,
where the latter relies on the ability of the $\fdia\gbox np$ hypothesis to `jump over' the subject
by means of $Comm_\lozenge$.

\begin{equation}\label{dutch}
\begin{array}{lll}
a. & \textrm{(ik weet dat) Bob$_{np}$ Alice$_{np}$ bewondert$_{np\bs(np\bs s)}$} &\\
b. & \textrm{(I know that) Bob$_{np}$ admires$_{(np\bs s)/np}$ Alice$_{np}$} & \\
c. & \textrm{man$_{n}$ die$_{??}$ de\_hond$_{np}$ bijt$_{np\bs(np\bs s)}$} & \\
d. & \textrm{man who bites the dog} & \textrm{(= subject relativization)}\\
e. & \textrm{man whom the dog bites} & \textrm{(= object relativization)}\\
\end{array}
\end{equation}

The crucial subderivations for the (\ref{dutch}c) example schematically rely on the following steps
(working upward): $\bs$ Introduction withdraws the $\fdia\gbox np$ hypothesis,
$\fdia$ Elimination followed by zero or more steps of structural reasoning bring the
hypothesis to the position where it can actually be used as a `regular' $np$, thanks
to the $\gbox$ Elimination proof of $\langle \gbox np\rangle\vdash np$.
The derived rule (\textit{xleft}) in (\ref{xleftfirst}) telescopes this sequence of
inference steps into a one-step inference, allowing for a succinct representation
of the derivations.

\begin{equation}\label{xleftfirst}
\begin{array}{ccc}
\scalebox{0.9}{
\hspace{-1cm}\infer[I\bs]{\nd{\Gamma[\Delta]}{\lambda^{l} x.{}^{c^n}t[{}^{\cup} x/z]:\fdia\gbox A\bs B}}{
\infer[E\>\fdia]{\nd{x:\fdia\gbox A\cdott\Gamma[\Delta]}{{}^{c^n}t[{}^{\cup} x/z]:B}}{
\infer[]{\nd{x:\fdia\gbox A}{x:\fdia\gbox A}}{}
\infer[(Ass_\fdia,Comm_\fdia)^n]{\nd{\langle z:\gbox A\rangle\cdott\Gamma[\Delta]}{{}^{c^n}t:B}}{
\infer[]{\vdots}{
\infer[]{\nd{\Gamma[\langle z:\gbox A\rangle\cdott\Delta]}{t:B}}{
\infer[]{\vdots}{
\infer[E\>\gbox]{\nd{\langle z:\gbox A\rangle}{{}^{\vee}z:A}}{
\infer[]{\nd{z:\gbox A}{z:\gbox A}}{}}}}}}}}  \quad 
\infer[\mathit{[xleft]^n}]{\nd{\Gamma[\Delta]}{\lambda^{l} x.
{}^{c^n}t[{}^\vee {}^{\cup} x/y]:\fdia\gbox A\bs B}}{\infer[]{\nd{\Gamma[y:A\cdott\Delta]}{t:B}}{\infer[]{\vdots}{\infer[]{[\nd{y:A}{y:A}]^n}{}}}}}
\end{array}
\end{equation} Here abbreviate the repeated application of the controlled commutativity rule on a single formula using the index $n$, where it serves a double purpose: indexing the hypothesis that will be extracted, and quantifying how many times the commutativity rule must be applied to licence this extraction. The proof term ${}^{c^n}t$ results from the $n$th application of this rule to the proof with conclusion term $t$, inductively defined with ${}^{c^0}t=t$ and ${}^{c^{n+1}}t={}^c({}c{^{n}}t)$.

Using our compiled inference rule, here are the derivations of both relativization readings, to be compared with those with the full uncompiled derivation in Appendix \ref{fulltrees}. On the proof of the subject relativization reading
(\ref{dutch}d), at the axioms, we show the constants (words) that will be substituted for
the parameters of the proof term for the derivation. Also, in the structure terms on the left of
the turnstile, we use these words instead of the parameter-type pairs to enhance legibility. 
This derivation uses the $\fdia\gbox np$ hypothesis as the subject of the relative clause
body; it simply relies on $\fdia$ and $\gbox$ Elimination, and doesn't involve structural
reasoning. 

\begin{equation*}
\infer[\bo \bs E \bc^{}]{\mbox{man}\cdot_{}(\mbox{die}\cdot_{}((\mbox{de}\cdot_{}\mbox{hond})\cdot_{}\mbox{bijt})) \vdash
\textcolor{red}{(y_{0}\triangleright (z_{0} \triangleleft \lambda^{l} x_{1}.{}^{c^0}({}^{\vee}  {}^{\cup} x_{1}\triangleright ((x_{2} \triangleleft  y_{2})\triangleright z_{2}))))} : n}{
      \infer[\ell]{\textcolor{red}{y_{0}} : n}{\mbox{man}}
   & 
      \infer[\bo / E \bc^{}]{\mbox{die}\cdot_{}((\mbox{de}\cdot_{}\mbox{hond})\cdot_{}\mbox{bijt}) \vdash 
\textcolor{red}{(z_{0} \triangleleft \lambda^{l} x_{1}.{}^{c^0}({}^{\vee}  {}^{\cup} x_{1}\triangleright ((x_{2} \triangleleft  y_{2})\triangleright z_{2})))} : n \bs_{}n}{
         \infer[\ell]{\textcolor{red}{z_{0}} : (n \bs_{}n) /_{}(\diamondsuit_{}\Box_{}np \bs_{}s)}{\mbox{die}}
      & 
         \infer[\bo \mathit{xleft} \bc^{0}]{(\mbox{de}\cdot_{}\mbox{hond})\cdot_{}\mbox{bijt} \vdash 
          \textcolor{red}{\lambda^{l} x_{1}.{}^{c^0}({}^{\vee}  {}^{\cup} x_{1}\triangleright ((x_{2} \triangleleft  y_{2})\triangleright z_{2}))}: \diamondsuit_{}\Box_{}np \bs_{}s}{
\infer[\bo \bs E \bc^{}]{\makebox[.85em]{\textvisiblespace}\cdot_{}((\mbox{de}\cdot_{}\mbox{hond})\cdot_{}\mbox{bijt}) \vdash 
\textcolor{red}{(x\triangleright ((x_{2} \triangleleft  y_{2})\triangleright z_{2}))} : s}{
      \bo \makebox[.85em]{\textvisiblespace} \vdash \textcolor{red}{x} : np \bc^{0}
   & 
      \infer[\bo \bs E \bc^{}]{(\mbox{de}\cdot_{}\mbox{hond})\cdot_{}\mbox{bijt} \vdash 
      \textcolor{red}{((x_{2} \triangleleft  y_{2})\triangleright z_{2})} : np \bs_{}s}{
         \infer[\bo / E \bc^{}]{\mbox{de}\cdot_{}\mbox{hond} \vdash \textcolor{red}{(x_{2} \triangleleft y_{2})} : np}{
            \infer[\ell]{\textcolor{red}{x_{2}} : np /_{}n}{\mbox{de}}
         & 
            \infer[\ell]{\textcolor{red}{y_{2}} : n}{\mbox{hond}}
         }
      & 
         \infer[\ell]{\textcolor{red}{z_{2}} : np \bs_{}(np \bs_{}s)}{\mbox{bijt}}
      }
   }}
      }
   }
\end{equation*} The index $0$ in the rule \textit{xleft} connects to the indexing of the hypothesis, reflecting that the hypothesis was already at the leftmost position. Therefore, no control rule is in need to be used. Contrast this with the derivation of the (\ref{dutch}e) object relativization interpretation. In
this case the $\fdia\gbox np$ hypothesis is manoeuvred to the direct object position in the
relative clause body thanks to the controlled commutativity option, used once as indicated by the index $1$ in the \textit{xleft} rule:

\begin{equation*}
\infer[\bo \bs E \bc^{}]{\mbox{man}\cdot_{}(\mbox{die}\cdot_{}((\mbox{de}\cdot_{}\mbox{hond})\cdot_{}\mbox{bijt})) \vdash
\textcolor{red}{(y_{0}\triangleright (z_{0} \triangleleft  \lambda^{l} x_{1}.{}^{c^1}((x_{2} \triangleleft y_{2})\triangleright({}^{\vee}{}^{\cup}x_{1}\triangleright z_{2}))))} : n}{
      \infer[\ell]{\textcolor{red}{y_{0}} : n}{\mbox{man}}
   & 
      \infer[\bo / E \bc^{}]{\mbox{die}\cdot_{}((\mbox{de}\cdot_{}\mbox{hond})\cdot_{}\mbox{bijt}) \vdash
 \textcolor{red}{(z_{0} \triangleleft  \lambda^{l} x_{1}.{}^{c^1}((x_{2} \triangleleft y_{2})\triangleright({}^{\vee}{}^{\cup}x_{1}\triangleright z_{2})))} : n \bs_{}n}{
         \infer[\ell]{\textcolor{red}{z_{0}} : (n \bs_{}n) /_{}(\diamondsuit_{}\Box_{}np \bs_{}s)}{\mbox{die}}
      & 
         \infer[\bo \mathit{xleft} \bc^{1}]{(\mbox{de}\cdot_{}\mbox{hond})\cdot_{}\mbox{bijt} \vdash
 \textcolor{red}{\lambda^{l} x_{1}.{}^{c^1}((x_{2} \triangleleft y_{2})\triangleright({}^{\vee}{}^{\cup}x_{1}\triangleright z_{2}))} : \diamondsuit_{}\Box_{}np \bs_{}s}{
         \infer[\bo \bs E \bc^{}]{(\mbox{de}\cdot_{}\mbox{hond})\cdot_{}(\makebox[.85em]{\textvisiblespace}\cdot_{}\mbox{bijt}) \vdash \textcolor{red}{(x_{2} \triangleleft y_{2})\triangleright(x\triangleright z_{2})} : s}{
      \infer[\bo / E \bc^{}]{\mbox{de}\cdot_{}\mbox{hond} \vdash \textcolor{red}{(x_{2} \triangleleft y_{2})} : np}{
         \infer[\ell]{\textcolor{red}{x_{2}} : np /_{}n}{\mbox{de}}
      & 
         \infer[\ell]{\textcolor{red}{y_{2}} : n}{\mbox{hond}}
      }
   & 
      \infer[\bo \bs E \bc^{}]{\makebox[.85em]{\textvisiblespace}\cdot_{}\mbox{bijt} \vdash \textcolor{red}{(x\triangleright z_{2})} : np \bs_{}s}{
         \bo \makebox[.85em]{\textvisiblespace} \vdash \textcolor{red}{x_{}} : np \bc^1
      & 
         \infer[\ell]{\textcolor{red}{z_{2}} : np \bs_{}(np \bs_{}s)}{\mbox{bijt}}
      }
   }}
      }
   }
\end{equation*}

Our aim in the following sections is to provide a compositional
interpretation of the control operators and the structural reasoning licensed by them that allows us to simultaneously represent the co-existing interpretations of ambiguous utterances such as (\ref{dutch}c).

\section{Interpretation Spaces}\label{interpretationspaces}
Let us turn to the action of the interpretation homomorphism on the \emph{types} of our extended Lambek calculus.
In the approach introduced in \cite{correia2020density}, types are sent to density matrix spaces. These spaces are set up in a directionality-sensitive way,  keeping in the semantics the distinction between left- or right-looking implications. Starting from the vector space $V$ and its dual $V^*$, we use a modified Dirac notation to distinguish between two sets of basis of $V$, $\left\{ \ket{_{i'}} \right\}$ and $\left\{\ket{^j}\right\}$, and two sets of basis of $V^*$,  $\left\{ \bra{^{j'}}\right\}$ and $\left\{ \bra{_i} \right\}$, obeying the orthogonality conditions $$\braket{_i}{_{i'}} = d_{ii'}, \quad \langle{^{j'}}\ket{^j} = d^{j'j}, \quad \braket{_i}{^j} = \delta_i^j, \quad \text{and} \quad  \braket{^j}{_i}=\delta_i^j,$$ where a metric function $d$ accounts for the eventual non-orthogonality between basis elements, and the Kronecker $\delta$ function defines the the relationship between dual basis elements. In general, the basis vector $\langle^{j'}|$ is obtained by the conjugate transposition of $\ket{^j}$. When the basis is not orthogonal, this operation does not render the dual basis vector of $\bra{^{j'}}$ (which by definition is orthogonal to it and in our notation is represented by $\ket{_{i'}}$), but another vector $\ket{^j}$ that requires the metric tensor to describe this relationship. Compare this with the case with only one set of basis for each space, obtained in the standard way: $\bra{^{j'}}$ coincides with $\ket{_{i'}}$ so that all basis vectors are orthogonal to each other, and the metric is just $\delta$

The \emph{basic} building block for the interpretations is the density matrix space $\tilde{V} \equiv V \otimes V^*$. This space has density matrices as elements, which we will use as the starting representations of words, instead of vectors. Density matrices are $1)$ positive operators with $2)$ trace normalized to $1$ \cite{nielsen2002quantum}. In a physical system, this means that we can not only access the quantum properties of states, expressed as a linear combination of basis states of $V$ or $V^*$, but we can also include the classical properties of a state, by constructing a basis of $\tilde{V}$ and describing the states as any linear combination formed with these basis elements that obeys conditions $1)$ and $2)$. Because the range of representations is enlarged, their use has been proposed for linguistic applications\cite{piedeleuopen,bankova2019graded,DBLP:journals/amai/SadrzadehKB18,coecke2020meaning}, which we expand on here focusing on including the directionality of the calculus in this distributional representation.
Defining the basis of $V$ and $V^*$ as we did before, we are able to construct a non-trivial basis for the density matrix space that carries over the structure of duality. For this space, we choose the basis formed by $\ket{_i}$ tensored with $\bra{_{i'}}$, $\tilde{E}=\left\{ \ket{_i}\bra{_{i'}} \right\} $. We define the dual density matrix space $\tilde{V}^* \equiv V \otimes V^*$ and assign to the dual basis of this space the map that takes each basis element of $\tilde{V}$ and returns a scalar. That basis is formed by $\bra{^j}$ tensored with $\ket{^{j'}}$, $\tilde{D}=\left\{ \ket{^{j'}}\bra{^j} \right\}$ , and is applied on the basis vectors of $\tilde{V}$ via the trace operation 

\begin{equation}
    \Tr \left(  \ket{_i}\braket{_{i'}}{^{j'}}\bra{^j}  \right) = \sum_l \braket{^l}{_i}\braket{_{i'}}{^{j'}}\braket{^j}{_l}.
\end{equation}

The \emph{composite} spaces are formed via the binary operation $\otimes$ (tensor product) and the unary operation $()^*$ (dual functor) 
that sends the elements of a density matrix basis to its dual basis, using the metric tensor. In the notation, we use $\tilde{A}$ for density matrix spaces (basic or compound),
and $\rho$, or subscripted $\rho_x, \rho_y, \rho_z,\ldots \in \tilde{A}$ for elements of such spaces. 
The $()^*$ operation is involutive; it interacts with the tensor product as $(\tilde{A} \otimes \tilde{B})^*= \tilde{B}^* \otimes \tilde{A}^*$  and acts as identity on matrix multiplication.

The homomorphism that sends syntactic types to semantic spaces is the map $\lceil . \rceil$.
For primitive types it acts as $$\lceil s \rceil = \tilde{S} \quad \text{and} \quad  \lceil np \rceil = \lceil n \rceil = \tilde{N},$$ with $S$ the vector space for sentence meanings, $N$ the space for nominal expressions (common nouns, full noun phrases).
For compound types we have 
$$    \lceil A/B \rceil= \lceil A \rceil \otimes \lceil B \rceil ^*  \quad \text{and} \quad  \lceil A \backslash B \rceil=  \lceil A\rceil^*  \otimes \lceil B\rceil. $$ This can be seen as an \textit{operational} interpretation of formulae: a dualizing functor acting on one of the types, followed by a tensor product, also a functor, are identified with particular operations on elements, specifically by multiplying with the elements of a metric or by taking the trace \footnote{Equivalentely, in a categorical distributional framework this corresponds to establishing a basis and taking either tensor contraction or multiplication as the operations that represent the $\eta$ and $\epsilon$ maps at the element level.}.

\subsection{Translation of unary modalities}
We now turn to how to send the formulae decorated with unary modalities to semantic spaces, in a way that stays in this functorial/operational framework.
Recall that in earlier work \cite{moortgat2017lexical, moortgat2019frobenius} modally
marked formulae are interpreted in the same space as their undecorated versions, i.e. $\lceil \lozenge A \rceil = \lceil \square A \rceil = \lceil A \rceil.$

To build a non-trivial interpretation of the unary connectives, we expand the interpretation space using the description of quantum states, distinguishing between their \emph{spatial} and \emph{spin} degrees of freedom. Let the $\lceil . \rceil$ homomorphism give a description of the \textit{spatial} components, encoding the numerically extracted distributional data.
In addition to the spatial component, and commuting freely with the spatial parts, we introduce a new vector space, a density matrix space $\mathfrak{S}$, with dimension $(N+1) \times (N+1)$, where $N$ the maximum value of index $n$ in the \textit{xleft} rule of eq.(\ref{xleftfirst}), where the \textit{spin} components are encoded. We denote this by the \textit{N-level spin space.} Here we do nftot distinguish between covariant and contravariant components, making the standard Dirac notation the appropriate one to deal with this space. Accordingly, the basis is orthonormal and has elements in $\left\{\ket{a}\bra{a'}\right\}$, with the values of $a$ and $a'$ ranging from $0$ to $N$. 

To obtain the full translation from syntactic types to their distributional interpretation spaces, we introduce an extended interpretation homomorphism that tensors the $\Spc{\cdot}$ interpretation of \emph{all} types with a density matrix space $\mathfrak{S}$ resulting in 

\begin{equation}
    \lfloor A \rfloor = \lceil A \rceil \otimes \mathfrak{S}.
\end{equation}

For atoms and slash types, $\Spc{\cdot}$ stays as defined.
For $\fdia A$ and $\gbox A$, we tensor $\Spc{A}$ with $\mathfrak{S}\otimes\mathfrak{S}^*$,
the type for the matrix representation of the operators associated with $\fdia$ and $\gbox$, that is, 
\begin{equation}
    \Spc{\fdia A} = \Spc{\gbox A} = \Spc{A}\otimes\mathfrak{S}\otimes\mathfrak{S}^*.
\end{equation} The key idea here is that by tensoring every type with an extra spin space via $\lfloor \cdot \rfloor$, the marked types have representations that encode maps from $\mathfrak{S}$ to $\mathfrak{S}$ coming from $\lceil \cdot \rceil$. This justifies the use of the same spin space to interpret the two markers, as they act as endomorphisms on the $\mathfrak{S}$ space coming from $\lceil \cdot \rceil$, as in for lozenge $\lfloor \lozenge A \rfloor = \lceil \lozenge A \rceil \otimes \mathfrak{S}$ and similarly for box. At the \emph{type} level, then, we find the structure to accommodate the operators $T_\fdia,T_{\gbox}\in\mathcal{L}(\mathfrak{S})$,
for which the concrete distinct interpretations will then be provided at the \emph{term} level. The key point of this structure is to give us precise control over the spin space as we interpret the unary modalities. Note that our connectives' interpretations do not interfere either with the distributional data that is stored in the spacial spaces, which is compatible with the interpretation of these connectives in previous work \cite{moortgat2017lexical, moortgat2019frobenius}. The interpretations we assign to the unary connectives consist of operations that only modify elements of an ancillary space. By enlarging the distributional space with this new spin space, we can effectivelly find a distributional meaning for the unary connectives.

As an example, here is the $\lfloor\cdot\rfloor$ mapping for
the relative pronoun type of (\ref{dutch}c).

\begin{align}
\Hm{(n\bs n)/(\fdia\gbox np\bs s)} &= \Spc{(n\bs n)/(\fdia\gbox np\bs s)}\otimes\mathfrak{S} \nonumber \\
 &=\Spc{n}^* \otimes \Spc{n} \otimes \Spc{s}^* \otimes \Spc{np} \otimes \underbrace{(\mathfrak{S}\otimes\mathfrak{S}^*)}_{T_{\lozenge}} \otimes 
 \underbrace{(\mathfrak{S}\otimes\mathfrak{S}^*)}_{T_{\square}} \otimes \mathfrak{S}\label{dietype}
\end{align}

\section{Operational Interpretation of Lambek Rules}\label{operationalinterpretation}

Given the new semantic spaces for the syntactic types, we now turn to the interpretation of the syntactic \emph{derivations}, as encoded by
their lambda proof terms, proving the soundness of the calculus presented in section \ref{types} with respect to the semantics of section \ref{interpretationspaces}.  In spin space, the operations that interpret different syntactic maps relate with the quantum postulates describing measurement and evolution of quantum systems\cite{nielsen2002quantum}. 

\paragraph{Quantum measurement:} Quantum measurements are described by a collection $M_a$ of
measurement operators, acting on the state space of the system being measured. The index $a$ refers to the measurement outcomes that
may occur in the experiment. If the state of the quantum system is $\rho$ immediately
before the measurement then the probability that result $a$ occurs is given by
$p(a) = \Tr(M_a^\dagger
M_a \rho)$
and the state of the system after the measurement is
\begin{equation}
    \rho_a := \frac{M_a \rho M^\dagger_a}{p(a)}.
\end{equation} The measurement operators satisfy the completeness equation, $\sum_a M^\dagger_a M_a = I$. For an observable $M$ with eigenvalues $m$ and eigenvectors $\ket{a}$, a \textit{projective} measurement is defined with $M_a= \ket{a}\bra{a}$; in this context we say that a state has been projected onto $\ket{a}\bra{a}$, and the quantum operator is then called a \textit{projector}. 
\paragraph{Evolution} The evolution of a closed quantum system is described by a unitary
transformation. That is, the state $\rho^i$ of the system at time $t_i$ is related to state
$\rho^{i+1}$ of the system at time $t_{i+1}$ by a unitary operator $U$ which depends only on these times. The state $\rho^{i+1}$ relates with the previous one $\rho^i$ by $\rho^{i+1}=U \rho^i U^\dagger$. 
\\

This correspondence is established via a function $\Dg{\cdot}$ that associates each term $t$ of type $A$ with
a semantic value, i.e.~an element of $\Spc{A}$, the semantic space where meanings of type $A$ live.
For proof terms, $\llbracket . \rrbracket$ is defined relative to an assignment function $g$, that provides a semantic value for
the basic building blocks, viz.~the variables that label the axiom leaves of a proof, in this case independently for the spatial ($S$) and spin ($\mathfrak{S}$) components. A particular assignment $g^S_{x,kk'}$ is used to interpret the lambda abstraction in the spatial spaces:

\begin{definition} \label{altg}
Given a variable $x$ of type $A$, we write $g^S_{x,kk’}$ for the
assignment exactly like $g^S$ except for the variable $x$, which takes the value of the basis element of the interpreting space $\ket{_{k}^{}} \prescript{}{\lceil A \rceil}{\bra{_{k'}}}$. 
\end{definition} 

The elements of the spin space are given by

\begin{equation}
\rho^\mathfrak{S}_x= \sum_{a,a'=0}^{n-1} {^\mathfrak{S}\textbf{X}_{aa'}} \ket{a} \prescript{}{\mathfrak{S}}{\bra{a'}}.
\end{equation}

A pair of special assignment functions $g^\mathfrak{S}_{x,I}$ and  $g^\mathfrak{S}_{x,y}$ is used to interpret the lambda abstraction in the spin space:

\begin{definition} \label{altgspin1}
Given a variable $x$ of type $A$, we write $g^\mathfrak{S}_{x,I}$ for the
assignment exactly like $g^\mathfrak{S}$ except for the variable $x$, which takes the value of the normalized identity, $I=\sum_a \frac{1}{\text{dim }{\mathfrak{S}}}\ket{a}  \prescript{}{\mathfrak{S}}{\bra{a}}$. 
\end{definition} 

\begin{definition} \label{altgspin2}
Given a variable $x$ of type $A$, we write $g^\mathfrak{S}_{x,y}$ for the
assignment exactly like $g^\mathfrak{S}$ except for the variable $x$, which takes the value of variable $y$, also of type $A$. 
\end{definition}

The spatial interpretation of terms of types formed with binary connectives is as given in \cite{correia2020density}. We reproduce here the main results, but focus on their interpretation in spin space. Further, we introduce the interpretation of the rules that introduce and eliminate unary connectives.

Some elimination rules will be interpreted in spin space using an instance of a projective measurement. Given a term $u$ of type A and another term $t$ of type $B$, we define a map $\llbracket t^A \rrbracket_{g^\mathfrak{S}} *\llbracket u^B \rrbracket_{g^\mathfrak{S}}: \mathfrak{S} \times \mathfrak{S} \rightarrow \mathfrak{S}$ acting on the interpretation of the terms in spin space:

\begin{equation}\label{conv}
\llbracket t^A \rrbracket_{g^\mathfrak{S}} * \llbracket u^B \rrbracket_{g^\mathfrak{S}} =  \frac{\left(\left\llbracket u^B \right\rrbracket_{g^\mathfrak{S}}\right)^\frac{1}{2} \cdot \left\llbracket t^{A} \right\rrbracket_{g^\mathfrak{S}} \cdot \left(\left\llbracket u^B \right\rrbracket_{g^\mathfrak{S}}\right)^\frac{1}{2}}{\Tr_\mathfrak{S} \left( \left(\left\llbracket u^B \right\rrbracket_{g^\mathfrak{S}}\right)^\frac{1}{2} \cdot \left\llbracket t^{A} \right\rrbracket_{g^\mathfrak{S}} \cdot \left(\left\llbracket u^B \right\rrbracket_{g^\mathfrak{S}}\right)^\frac{1}{2} \right)},
\end{equation} with $(.)^{\frac{1}{2}}$ such that when applied on an operator $R$ we have that $(R)^{\frac{1}{2}} \cdot (R)^{\frac{1}{2}}= R$. Positive operators, such as density matrices, have a unique positive square root \cite{axler1997linear}. Physically, the spin split in its square-root acts as a measurement operator on the other input spin. Using normalization, the outcome is a well defined spin state. An unnormalized version of this operator is defined in \footnote{This a generalization of one of the Frobenius algebras already used in \cite{bankova2019graded} in the category \textbf{CPM(FHilb)}, where, given the full density matrix representations of sentence, noun and verb, respectively $\rho(s)$, $\rho(n)$ and $\rho$, they relate by $\rho(s)=\rho(n)^{\frac{1}{2}} \rho(v) \rho(n)^{\frac{1}{2}}$.}. An unnormalized version of this map is defined as the "phaser" in \textit{Coecke and Meichanetzidis}\cite{coecke2020meaning}.

\subsection{Axiom}

The axiom will be given by an element of the spatial spaces, tensored with an element of the spin space.

\begin{equation}
  \left\llbracket x^A \right\rrbracket_g = g(x^A) = \prescript{}{}{\rho_x^{\lfloor A \rfloor}}=  \left\llbracket x^A \right\rrbracket_{g^S}  \otimes  \left\llbracket x^A \right\rrbracket_{g^\mathfrak{S}}, 
\end{equation} where

\begin{equation} \label{axiom}
\quad
     \left\llbracket x^A \right\rrbracket_{g^\mathfrak{S}} = \sum_{aa'} {^\mathfrak{S}\textbf{X}_{aa'}} \ket{a} \prescript{}{\mathfrak{S}}{\bra{a'}}
\quad\textrm{and}\quad
     \left\llbracket x^A \right\rrbracket_{g^S} =  \sum_{ii'} \prescript{}{}{^S\textbf{X}^{ii'}}\ket{_i} \prescript{}{\lceil A \rceil}{\bra{_{i'}}}. 
\end{equation}

\subsection{Introduction and elimination of binary connectives}

\paragraph{Elimination of $\slash$ and $\backslash$} \mbox{}


\begin{align}\label{triangleleft}
& 
\left\llbracket (t\triangleleft u)^B \right\rrbracket_g   \equiv \Tr_{\lceil A \rceil} \left(\left\llbracket t^{B\slash A} \right\rrbracket_{g^S} \cdot  \left\llbracket u^A \right\rrbracket_{g^S} \right) \otimes \llbracket t^{B\slash A} \rrbracket_{g^\mathfrak{S}} * \llbracket u^A \rrbracket_{g^\mathfrak{S}}.
\end{align} 


\begin{align}\label{triangleright}
& 
\left\llbracket (u\triangleright t)^B \right\rrbracket_{g} \equiv \Tr_{\lceil A \rceil} \left( \left\llbracket u^A \right\rrbracket_{g^S} \cdot \left\llbracket t^{A\backslash B} \right\rrbracket_{g^S}  \right) \otimes\llbracket t^{A\backslash B} \rrbracket_{g^\mathfrak{S}} * \llbracket u^A \rrbracket_{g^\mathfrak{S}}.
\end{align} 

\paragraph{Introduction of $\slash$ and $\backslash$}\mbox{} 


\begin{align}\label{lambdar}
& 
\left\llbracket \left( \lambda^r x. t \right)^{B\slash A} \right\rrbracket_{g} \equiv \sum_{kk'} \left(\llbracket t^B \rrbracket_{g^{S}_{x,kk'}} \otimes \ket{^{k'}_{}} \prescript{}{\lceil A \rceil^*}{\bra{^{k}}} \right) \otimes \left\llbracket t^{B} \right\rrbracket_{g_{x,I}^\mathfrak{S}}.
\end{align}


\begin{align}\label{lambdal}
& 
\left\llbracket \left( \lambda^l x. t \right)^{A\bs B} \right\rrbracket_{g} \equiv \sum_{kk'} \left( \ket{^{k'}_{}} \prescript{}{\lceil A \rceil^*}{\bra{^{k}}} \otimes \llbracket t^B \rrbracket_{g^{S}_{x,kk'}}   \right) \otimes \left\llbracket t^{B} \right\rrbracket_{g_{x,I}^\mathfrak{S}}.
\end{align}

Syntactic equalities like beta reduction are interpreted as equalities in this model, as is shown in appendix \ref{betareduction}. 

\subsection{Introduction and elimination of unary connectives}

As seen earlier in the example of eq.(\ref{dietype}), at the term level the diamond introduction is interpreted by the map $T_\lozenge$ and box introduction is interpreted by the map $T_\square$, both consisting of maps $\mathfrak{S} \rightarrow \mathfrak{S}$. Two more operations need to be introduced, namely those that eliminate box, $T'_\square$, and that eliminate diamond $T'\lozenge$. Since these are the maps applied in our proof, we next give their explicit form.

The operation $T'_\square$ acting on elements of $\mathfrak{S}$ is the linear combination of projectors $T'^a_\square$ onto pure states used as projectors $M_a=\ket{a} \prescript{}{\mathfrak{S}}{\bra{a}}$, generated by the eigenstates of an observable with $N+1$ different eigenvalues, specified for a particular unary modality, indexed by $a \in \{0,...,N\}$. Applied on a state $\rho^\mathfrak{S}_x$, the general result is the following mixed state 

\begin{align}\label{tdiamond}
&T'_\square (\rho^\mathfrak{S}_x) = \sum_{a=0}^{N} c_a T'^a_\square (\rho^\mathfrak{S}_x) \equiv \sum_{a=0}^{N} c_a (\rho^\mathfrak{S}_x * \ket{a} \prescript{}{\mathfrak{S}}{\bra{a}}) = \sum_{a=0}^{N} c_a \left( \frac{M_a \rho^\mathfrak{S}_x M_a}{\Tr \left(M_a \rho^\mathfrak{S}_x M_a \right)} \right),
\end{align} with $\sum_{a=0}^N c_a =1$, $c_a \in R$. Defining the ordering of the eigenstates by the increasing value of their corresponding index $a$, rule $E_\square$ will be interpreted in the spin components as the projection onto the lowest eigenstate, effectively with $c_0=1$ and $c_{a\neq 0}=0$. 

The operation $T'_\lozenge$ acts on elements by performing a unitary transformation, generated by the successive application of matrices $U_0=\mathbb{1}$ and $U_b\in SU(N+1)$ on density matrices, for $b\in \{ 1,\dots,N^2+2N\}$, represented as $T'^b_\lozenge$, for a particular representation and ordering. Again applied to the state $\rho^\mathfrak{S}_x$, the application of this operation is

\begin{align} \label{tsquare}
\left(T'^b_\lozenge \left(\rho^\mathfrak{S}_x\right) \right)^{d_b}= \left\{\begin{matrix}
&\rho^\mathfrak{S}_x \; & \text{if} \; d_b=0\\ 
&U_b \rho^\mathfrak{S}_x U_b^\dagger \; & \text{if} \; d_b=1
\end{matrix}\right. 
\end{align}
\begin{align}\label{recursive}
T'_\lozenge( \rho^\mathfrak{S}_x ) = \left( T'^{N^2+2N}_\lozenge\left( T'^{N^2+2N-1}_\lozenge\left(...  \left(T'^0_\lozenge\left(\rho^\mathfrak{S}_x\right) \right)^{d_0}\right)\right)^{d_{N^2+2N-1}}\right)^{d_{N^2+2N}}
\end{align} where $()^\dagger$ indicates hermitian conjugation and $d_b\in \{0,1\}$ \footnote{Eq. (\ref{recursive}) can possibly be extended with permutations over the order of application of $T'^b_\lozenge$.}. The rule $E_\lozenge$ is thus interpreted as performing a unitary transformation, using that $d_0=1$ and $d_{b\neq 0} \neq 0$.

In the particular case where we interpret the introduction of a connective with the same operation of its connective, that is $T_\square = T'_\lozenge$ and $T_\lozenge=T'_\square$, the adjoint properties of the unary connectives are preserved. The implications $\lozenge \square A \rightarrow A \rightarrow \square \lozenge A$ are interpreted on space $\mathfrak{S}$  as $$T_\lozenge\left(T_\square \left(\mathfrak{S} \right) \right) \in \mathfrak{S} \in T_\square\left(T_\lozenge \left(\mathfrak{S} \right) \right).
$$ In the first inclusion we have a unitary transformation followed by a projection, which is inside the interpretation space of the state, the entire Bloch sphere. For second inclusion, any state inside of the Bloch sphere is inside the scope of projections followed by a unitary transformation. This is a consequence of the non-commutativity of the operations that interpret these connectives, measurement and evolution.

\paragraph{Elimination of $\square$:} $\llbracket (^\vee t)^B \rrbracket_g = \llbracket t^{\square B} \rrbracket_{g^S} \otimes T'^0_{\square} \left(\llbracket t^{\square B} \rrbracket_{g^{\mathfrak{S}}}\right)$


\paragraph{Elimination of $\lozenge$:}\mbox{} 

%
%
\begin{align}
    \llbracket ^\cup t \rrbracket_{g^\mathfrak{S}} = T'^0_\lozenge \left(\llbracket t^{\lozenge A} \rrbracket_{g^\mathfrak{S}}\right)
\end{align}

\begin{align}
&\left\llbracket (u[^\cup t \slash x])^B \right\rrbracket_g = \Tr_{\lceil A \rceil} \left( \left\llbracket t^{\lozenge A} \right\rrbracket_{g^S} \cdot \sum_{kk'}   \ket{^{k'}_{}} \prescript{}{\lceil A \rceil^*}{\bra{^{k}}} \otimes \llbracket u^B \rrbracket_{g^{S}_{x,kk'}} \right) \otimes \llbracket u^B \rrbracket_{g^\mathfrak{S}_{x,^\cup t}}. 
\end{align}

\paragraph{Introduction of $\square$ and $\lozenge$:}\mbox{}  


\begin{equation}
\left\llbracket (^\wedge t)^{\square B} \right\rrbracket_g = \left\llbracket  t^{B} \right\rrbracket_{g^S} \otimes  T^0_\square\left(\llbracket t^{B} \rrbracket_{g^\mathfrak{S}}\right), \quad
\left\llbracket (^\cap t)^{\lozenge B} \right\rrbracket_g = \llbracket t^B \rrbracket_{g^S} \otimes T^0_\lozenge \left(\llbracket t ^B \rrbracket_{g^{\mathfrak{S}}}\right)
\end{equation}

\subsection{Structural Reasoning}\label{structural}

To interpret the derived infererence rule, a \textit{raising operator} $S_+$ acts on the input state and is applied as many times as nodes that need to be jumped to be in the right position to be extracted. We record that information by an index $m$ on the substitution brackets of the proof term encoding the (\emph{xleft}) inference. The index acts as a power on the raising operator, $(S_+)^m$, changing a state $\rho_a=\ket{a} \prescript{}{\mathfrak{S}}{\bra{a}}$ to $\rho_{a+m}=\ket{a+m} \prescript{}{\mathfrak{S}}{\bra{a+m}}$, where we use the convention that a matrix to the zeroth power is the identity matrix. Note that this is not a unitary operator, which means that the resulting state must be normalized after the application. Additionally the derived inference rule is interpreted using the previously given interpretations of $\square$ and $\lozenge$.



\paragraph{Derived Inference Rule} \mbox{}

$\mathit{[xleft]^n}$: Premise $t^B$ with subterm $y^A$ at location $n$; conclusion $(\lambda^{l} x.\left({}^{c^n}t\right)^B[{}^\vee {}^{\cup} x/y]^{n})^{\fdia\gbox A\bs B}$: 


\begin{align} \label{projuni}
    \left\llbracket \left({}^\vee {}^{\cup} x \right)^A \right\rrbracket_{g^{\mathfrak{S}}} =T^0_\lozenge \left(  T^0_\square \left( \left\llbracket  x^{\lozenge \square A} \right\rrbracket_{g^{\mathfrak{S}}} \right) \right)
\end{align}

\begin{align}\label{xleftprev}
 & \llbracket (\lambda^{l} x.\left({}^{c^n}t\right)^B[{}^\vee {}^{\cup} x/y])^{\fdia\gbox A\bs B} \rrbracket_g = \nonumber \\
  = & \sum_{ll'} \ket{^{l'}_{}} \prescript{}{\lceil A \rceil^*}{\bra{^{l}}} \otimes \left[\Tr_{\lceil A \rceil} \left( \left\llbracket  x^{\lozenge \square A} \right\rrbracket_{g^{S}} \cdot \sum_{kk'}   \ket{^{k'}_{}} \prescript{}{\lceil A \rceil^*}{\bra{^{k}}} \otimes \llbracket t^B \rrbracket_{g^{S}_{y,kk'}} \right) \right]_{g^{S}_{x,ll'}} \nonumber \\
  & \otimes \frac{\left[ \left(S_+\right)^n
  \llbracket t^B \rrbracket_{g^{\mathfrak{S}}_{y,{}^\vee {}^{\cup} x}} \left(\left(S_+\right)^\dagger\right)^n   \right]_{g^{\mathfrak{S}}_{x,I}}}{\Tr_\mathfrak{S} \left(  \left[ \left(S_+\right)^n  \llbracket t^B \rrbracket_{g^{\mathfrak{S}}_{y,{}^\vee {}^{\cup} x}} \left(\left(S_+\right)^\dagger\right)^n   \right]_{g^{\mathfrak{S}}_{x,I}} \right)}
\end{align}

Here we can see clearly the physical meaning that the quantum interpretation gives to the application of the modal operators. In eq.(\ref{projuni}), the combination of application of $T'^0_\lozenge$ and $T'^0_\square$, interpreted as a projection and a unitary operation, respectively, takes the form of one of the possible outcomes of the quantum process $E=PU$ \cite{nielsen2002quantum}, applied on the state $\left\llbracket  x^{\lozenge \square A} \right\rrbracket_{g^{\mathfrak{S}}}$, namely the one where the final state is $\bra{0}\prescript{}{\mathfrak{S}}{\ket{0}}$. Having the unary connectives interpreted with the non-commutative operations of projection and unitary transformation correctly preserves the order of application of the connectives imposed at the syntactic level. The derivation of this interpretation from the extended version of \textit{xleft} rule is explored in Appendix \ref{xleftinterp}.

\section{Two-level spin space}\label{spinspace}

\begin{figure}[h]
\centering
\includegraphics[scale=0.07]{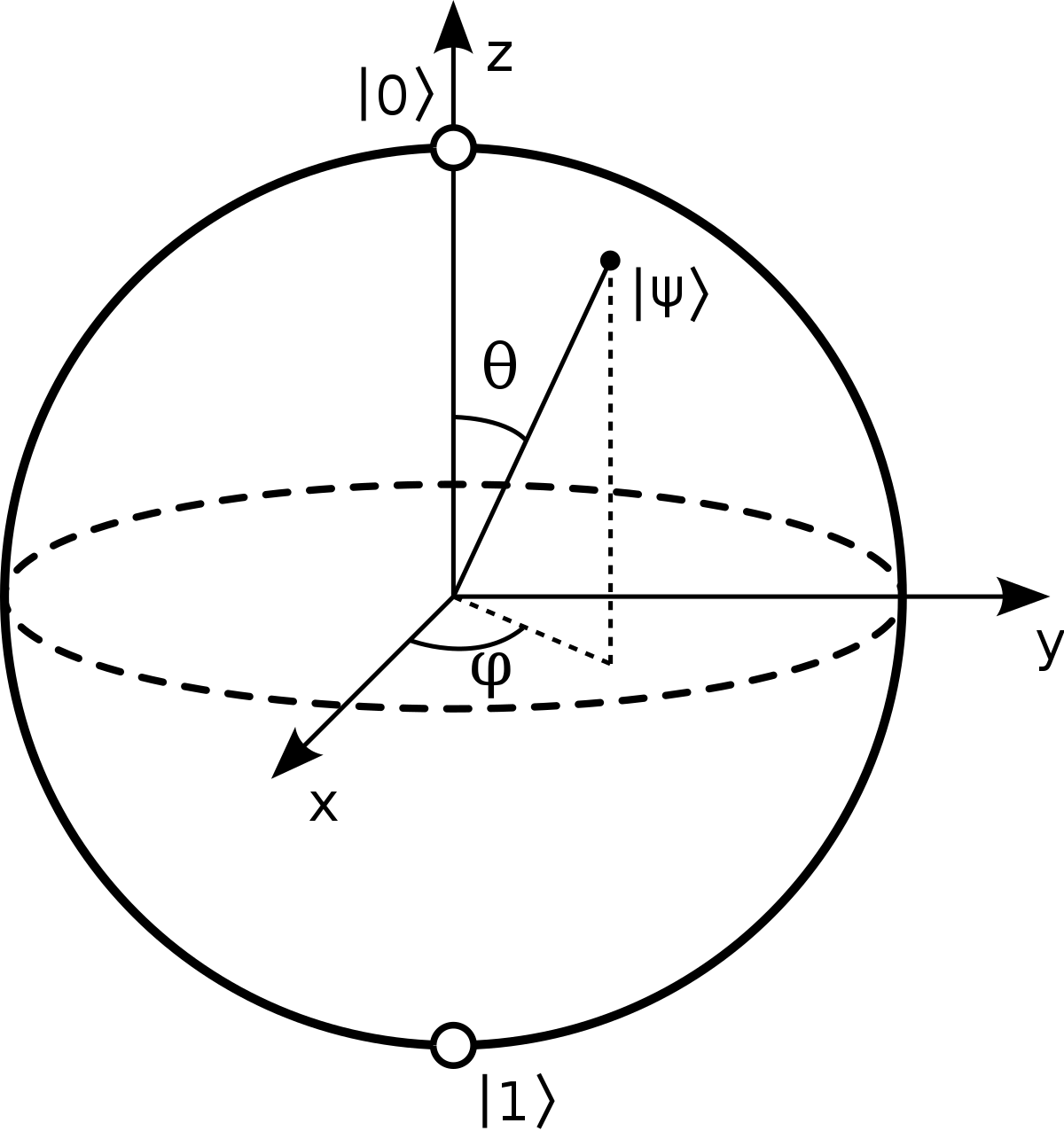}
\caption{\small Bloch sphere representation of a two-level quantum state, also called a \textit{qubit}. The general form of a state on the surface is $\ket{\Psi}=\left(\cos \frac{\theta}{2} \ket{0} + e^{i\phi} \sin\frac{\theta}{2} \ket{1}\right)e^{i\gamma}$. The global phase $e^{i\gamma}$ is not represented because it has no effect on the density matrix.  A product of states $^{pure}\rho^\mathfrak{S}= \ket{\Psi} \prescript{}{\mathfrak{S}}{\bra{\Psi}}$ is called a \textit{pure state}, represented on the surface of the sphere. Otherwise the states are called \textit{mixed states} and live inside of the sphere.}
\label{bloch}
\end{figure}

The structural ambiguity at hand will be treated using a two-level spin space, since we have two ambiguous readings. This space is used to encode spin states of fermionic particles, with spin $1\slash2$, such as electrons and protons. A helpful geometric visualization of the states in this space is the $\textit{Bloch sphere}$, in fig. \ref{bloch}.

To interpret the action of the unary connectives in the spin space, we suppose that the particles with spin, our words in this case, are subjected to a uniform magnetic field pointing in the $z$ direction. Using natural units, let 
$$S_z=\frac{1}{2}\begin{pmatrix}
1 & 0\\ 
0 & -1
\end{pmatrix}$$ be the spin operator in the $z$ direction. The eigenvectors of this operator are the orthogonal states $\ket{0}=(0,1)^\intercal$ and $\ket{1}=(1,0)^\intercal$, using the standard matrix representation. On the Bloch sphere, these states correspond to the north and south poles, respectively. The corresponding eigenvalues are $e_0=-1/2$ and $e_1= 1/2$. This is the operator that we will use to interpret our unary modality. Thus $T_\lozenge$ is the set formed by linear combinations of states $\rho_0=\ket{0} \prescript{}{\mathfrak{S}}{\bra{0}}$ and $\rho_1=\ket{1} \prescript{}{\mathfrak{S}}{\bra{1}}$, the states that lie on the $z$-axis inside the Bloch sphere.

To interpret controlled commutativity, we use the raising operator is $$S_+= S_x + iS_y = \begin{pmatrix}
0 & 1\\ 
0 & 0
\end{pmatrix}.$$ Once applied on $\rho_0$ the result is $\rho_1$, and a further application has a null result. Note that, together with with the lowering operator $$S_-= S_x - iS_y = \begin{pmatrix}
0 & 0\\ 
1 & 0
\end{pmatrix},$$ it obeys the completeness relation $S_+ \left(S_+\right)^\dagger + S_- \left(S_-\right)^\dagger = \mathbb{1}$.

\section{Going Dutch again}\label{dutchrelclauses}

To illustrate the interpretation process, we return to our Dutch relative clause example "man die de hond bijt", and show how we handle the derivational ambiguity.
The lexicon below has the syntactic type assignments and the corresponding semantic spaces:

\[\begin{array}{r|c|l}
 & \textrm{syn type $A$} & \lfloor A\rfloor\\\hline
\textrm{man'} & n &  \tilde{N} \otimes \mathfrak{S},\\
\textrm{die'} & (n\backslash n)/(\lozenge \square np \backslash s) & \tilde{N}^* \otimes \tilde{N} \otimes \tilde{S}^* \otimes \tilde{N} \otimes (\mathfrak{S}\otimes\mathfrak{S}^*) \otimes (\mathfrak{S}\otimes\mathfrak{S}^*) \otimes \mathfrak{S}, \\
\textrm{de hond'} & np & \tilde{N} \otimes \mathfrak{S}, \\
\textrm{bijt'} & np \backslash np \backslash s & \tilde{N}^* \otimes \tilde{N}^* \otimes \tilde{S} \otimes \mathfrak{S}.\\
\end{array}\] In order to compute the interpretations given by the two above derivations, we start from the following primitive interpretations:

\begin{align}
&\llbracket \text{man'}^n \rrbracket_I = \sum_{rr',ii'} {^S\textbf{M}}^{rr'} \ket{_{r}}\prescript{}{\lceil N \rceil}{\bra{_{r'}}} \otimes ^{\mathfrak{S}}\textbf{M}_{ii'} \ket{i}\prescript{}{\mathfrak{S}}{\bra{i'}}, \label{man} \\
&\llbracket \text{die'}^{(n\backslash n)/(\lozenge \square np \backslash s)} \rrbracket_I = \sum_{kk', ll', mm', nn', ii'} {^S\textbf{D}_{k'k\;m'm}^{\; \; \; ll'\; \; \;nn'}} \ket{^{k'\;m'}_{\; \; \; l\; \; \;n} } \prescript{}{\lceil N \rceil^* \otimes \lceil N \rceil \otimes (\lceil S \rceil^* \otimes \lceil N \rceil)}{\bra{^{k\;m}_{\; \;l'\; \;n'\;}}} \otimes ^\mathfrak{S} \textbf{D}_{ii} \ket{i}\prescript{}{\mathfrak{S}}{\bra{i}}; \label{die} \\
&\llbracket \text{de hond'}^{np} \rrbracket_I = \sum_{jj', ii'} {^S \textbf{H}^{jj'}} \ket{_{j}}\prescript{}{\tilde{N}}{\bra{_{j'}}} \otimes ^\mathfrak{S} \textbf{H}_{ii'} \ket{i}\prescript{}{\mathfrak{S}}{\bra{i'}}; \label{hond} \\
&\llbracket \text{bijt'}^{np\backslash np \backslash s} \rrbracket_I = \sum_{oo', pp', qq', ii'} {^S\textbf{B}_{o'o, p'p}^{\;\;\; \; \;\; \; \;\;qq'}} \ket{^{o'p'}_{\;\;\; \; \;\;q}} \prescript{}{\lceil N \rceil^* \otimes \lceil N \rceil^* \otimes \lceil S \rceil}{\bra{^{o p}_{\;\;\; \; q'}}} \otimes ^\mathfrak{S}\textbf{B}_{ii'} \ket{i}\prescript{}{\mathfrak{S}}{\bra{i'}}. \label{bijt}
\end{align} To obtain the correct contractions in the spatial components, that are related either to the subject or object relativization readings, the role of the hypothesis $x$ is crucial: interpreted as in eq.(\ref{axiom}), it contracts with the interpretation of "bijt" as the interpretations of the slash elimination rules prescribe, either in subject or object position. Its most important role is in the latter, blocking "de hond" from taking the immediate object position contraction. After that, variable $x$ is extracted using the \textit{xleft} rule, in a way that keeps all the other contractions unchanged, and keeping the right form such that "die" can contract in the correct position. This process is worked out in Appendix \ref{distinter}.

With respect to the spin components, the goal is that a pure state is preserved as it interacts with other spin states via slash elimination. As the hypothesis of type $\lozenge \square np$ is abstracted over, it attains the value of the identity matrix, onto which the box and diamond eliminations are applied, projecting it to the $\rho_0$ state. If the controlled commutativity rule is is applied, the raising operator brings this pure state to the orthogonal pure state $\rho_1$. In this way, each of the two readings is stored in one of orthogonal eigenstates of the $S_z$ operator, which are necessarily pure states. As they interact with "man" using the $(.)*(.)$ map, we predict that the final spin states will remain pure, using the result of Lemma 4.1 on the phaser in \textit{Coecke and Meichanetzidis}\cite{coecke2020meaning}, since the spin state that represents "man" interacts with a pure state in argument position. The full calculations are shown in Appendix \ref{spininter}.

The relative clause of the first reading has the interpretation

\begin{equation} \label{subrel}
\llbracket \text{die\_de\_hond\_bijt'} \rrbracket^1_I =  \sum_{rr',ll',jj',mm',nn'}  {^S\textbf{D}_{r'r\;m'm}^{\; \; \; ll' \;nn'}} \; {^S\textbf{H}^{jj'}} \; {^S\textbf{B}_{j'j, n'n}^{\;\;\; \; \;\; \; \;\;mm'}} \ket{^{r'} _{\;\;l}} \prescript{}{\lceil N \rceil}{\bra{^{r}_{\;\; l'}}} \otimes \ket{0} \prescript{}{\mathfrak{S}}{\bra{0}},
\end{equation} while for the second reading the interpretation is it is

\begin{equation} \label{objrel}
\llbracket \text{die\_de\_hond\_bijt'} \rrbracket^2_I =  \sum_{rr',ll',jj'mm',nn'} {^S\textbf{D}_{r'r\;m'm}^{\; \; \; ll' \;nn'}} \; {^S\textbf{H}^{jj'}} \; {^S\textbf{B}_{n'n,j'j}^{\;\;\; \; \;\; \; \;\;mm'}} \ket{^{r'}_{l}} \prescript{}{\lceil N \rceil}{\bra{^{r}_{l'}}} \otimes \ket{1} \prescript{}{\mathfrak{S}}{\bra{1}}.
\end{equation}

\begin{figure}[h]
\centering
\includegraphics[scale=0.2]{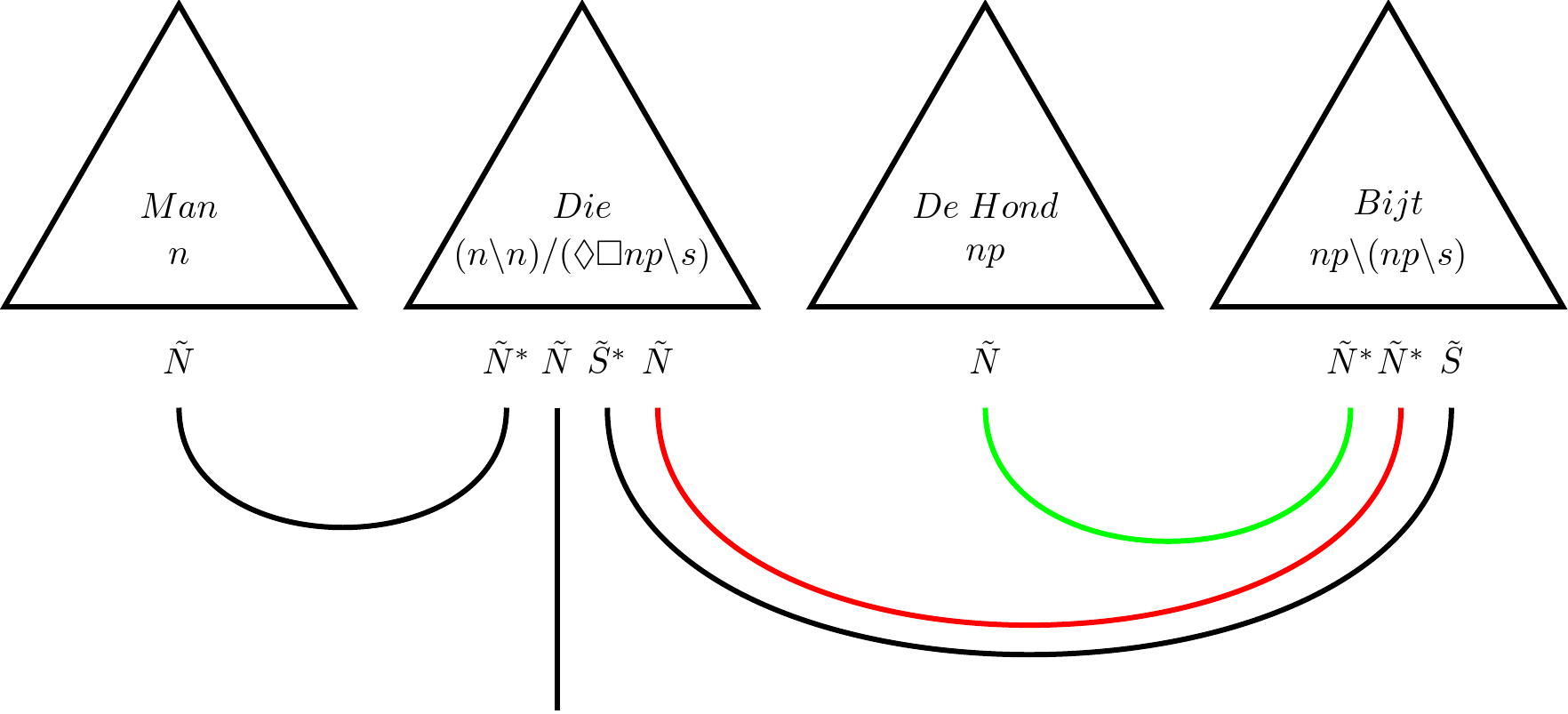}
\caption{\small Representation of  spatial contractions corresponding to the subject relativization reading of "man die de hond bijt", according to eq.(\ref{subrel}).}
\label{1stinter}
\end{figure}

\begin{figure}[h]
\centering
\includegraphics[scale=0.2]{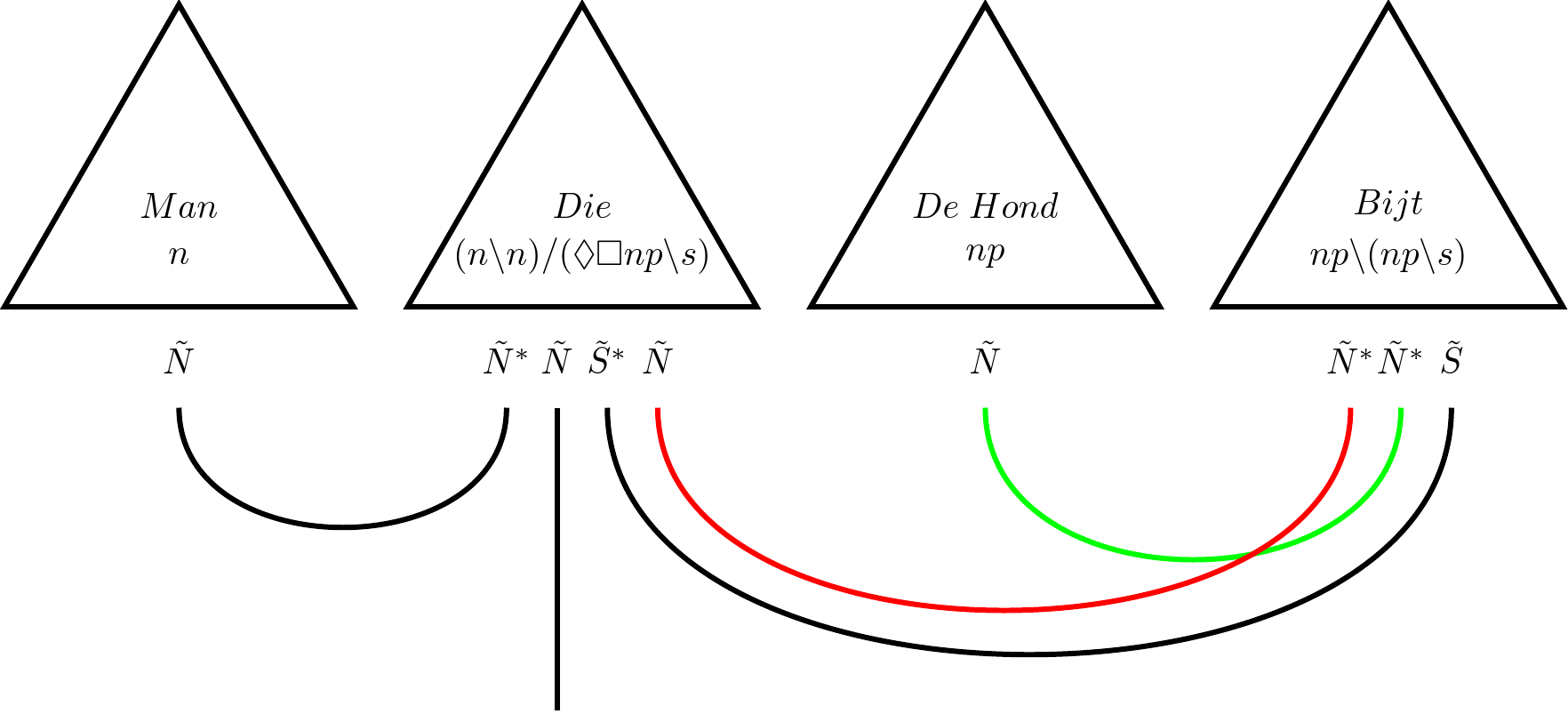}
\caption{\small Representation of contractions corresponding to the object relativization reading of "man die de hond bijt", according to eq.(\ref{objrel}).} 
\label{2ndinter} 
\end{figure} 

The final interpretation of the ambiguous phrase is given by the direct sum of the two unambiguous interpretations, weighted by parameters $p_1$ and $p_2$ that express the likelihood of each reading:

\begin{equation}\label{final}
\llbracket \mathit{man \_ die\_ de\_ hond\_ bijt} \rrbracket_I = p_1 \llbracket \mathit{man\_die\_de\_hond\_bijt} \rrbracket^1_I \oplus p_2 \llbracket \mathit{man\_die\_de\_hond\_bijt} \rrbracket^2_I.
\end{equation}

\section{Discussion and Conclusion}\label{conclusion}

In this paper we extended the interpretation space with a spin degree of freedom, showing how that can preserve extra information about the proof. We showed how interpreting the meanings of words directly as density matrices introduces a framework that can be used to encode higher-level content. This was done by interpreting the unary connectives as quantum operations in the spin space, such the information about the readings is preserved via a quantum process. When more than two ambiguous readings are possible, it constitutes future work to show that our framework can be extended by using a larger spin space and an appropriate raising operator. Besides its usefulness to deal with ambiguity, in future work we want also to study how the spin degree of freedom is suitable to distinguish the representations of marked types in a multimodal setting, possibly by associating them with eigenstates of different operators. While in this work the spin degree of freedom plays no bigger role than an extra two-dimensional degree of freedom, when going to a multimodal setting the interactions between the different spin eigenvectors will have quantum properties due to the non-commutativity of the operators. Interesting too is to relate our approach, where lambda terms are directly interpreted using elements and operations over them, with Kripke frames on vector spaces \cite{greco2019vector}, defining the valuation sets with the accessibility relations that translate into our operations, unveiling a stronger connection with the logic of residuation. Also relevant would be to compare our take on interpreting certain logic connectives using quantum mechanical operations with the mirror field of quantum logic \cite{coecke2000operational} that aims at interpreting quantum mechanics using logic tool, particularly modal logic \cite{coecke2000operational} which is at the root of our unary connectives, where too an association between projections and the logic of possibility ($\lozenge$ in our notation) is suggested. Finally, further research will have to show how the probability coefficients can be extracted from derivational data, and whether it is possible to go from the subject relativization reading to the object relativization reading applying only permutation operators as is done in \cite{correia2020density} for syntactic ambiguities and, in that case, what is precisely the connection with the derivation. Other interesting questions relate to finding the appropriate categorical interpretation of the spin space and operations that take place there, further helping us to relate our interpretation of control modalities with other logical operators that are also syntactic but do affect the meaning of a sentence, such as negation or quantification, but these are outside the scope of the present paper.



\nocite{*}
\bibliographystyle{eptcs}
\bibliography{generic}

\begin{thebibliography}{10}
\providecommand{\bibitemdeclare}[2]{}
\providecommand{\surnamestart}{}
\providecommand{\surnameend}{}
\providecommand{\urlprefix}{Available at }
\providecommand{\url}[1]{\texttt{#1}}
\providecommand{\href}[2]{\texttt{#2}}
\providecommand{\urlalt}[2]{\href{#1}{#2}}
\providecommand{\doi}[1]{doi:\urlalt{http://dx.doi.org/#1}{#1}}
\providecommand{\bibinfo}[2]{#2}

\bibitemdeclare{book}{axler1997linear}
\bibitem{axler1997linear}
\bibinfo{author}{Sheldon \surnamestart Axler\surnameend}
  (\bibinfo{year}{1997}): \emph{\bibinfo{title}{Linear algebra done right}}.
\newblock \bibinfo{publisher}{Springer Science \& Business Media},
   \doi{10.1007/b97662}.

\bibitemdeclare{article}{bankova2019graded}
\bibitem{bankova2019graded}
\bibinfo{author}{Dea \surnamestart Bankova\surnameend}, \bibinfo{author}{Bob
  \surnamestart Coecke\surnameend}, \bibinfo{author}{Martha \surnamestart
  Lewis\surnameend} \& \bibinfo{author}{Dan \surnamestart Marsden\surnameend}
  (\bibinfo{year}{2019}): \emph{\bibinfo{title}{Graded hyponymy for
  compositional distributional semantics}}.
\newblock {\sl \bibinfo{journal}{Journal of Language Modelling}}
  \bibinfo{volume}{6}(\bibinfo{number}{2}), pp. \bibinfo{pages}{225--260},
   \doi{10.15398/jlm.v6i2.230}.

\bibitemdeclare{article}{coecke2020meaning}
\bibitem{coecke2020meaning}
\bibinfo{author}{Bob \surnamestart Coecke\surnameend} \&
  \bibinfo{author}{Konstantinos \surnamestart Meichanetzidis\surnameend}
  (\bibinfo{year}{2020}): \emph{\bibinfo{title}{Meaning updating of density
  matrices}}.
\newblock {\sl \bibinfo{journal}{Journal of Applied Logics}} \bibinfo{volume}{7}(\bibinfo{number}{5}).


\bibitemdeclare{incollection}{coecke2000operational}
\bibitem{coecke2000operational}
\bibinfo{author}{Bob \surnamestart Coecke\surnameend}, \bibinfo{author}{David
  \surnamestart Moore\surnameend} \& \bibinfo{author}{Alexander \surnamestart
  Wilce\surnameend} (\bibinfo{year}{2000}): \emph{\bibinfo{title}{Operational
  quantum logic: An overview}}.
\newblock In: {\sl \bibinfo{booktitle}{Current research in operational quantum
  logic}}, \bibinfo{publisher}{Springer}, pp. \bibinfo{pages}{1--36},
\doi{10.1007/978-94-017-1201-9}.

\bibitemdeclare{article}{coecke2010mathematical}
\bibitem{coecke2010mathematical}
\bibinfo{author}{Bob \surnamestart Coecke\surnameend},
  \bibinfo{author}{Mehrnoosh \surnamestart Sadrzadeh\surnameend} \&
  \bibinfo{author}{Stephen \surnamestart Clark\surnameend}
  (\bibinfo{year}{2010}): \emph{\bibinfo{title}{Mathematical foundations for a
  compositional distributional model of meaning}}.
\newblock {\sl \bibinfo{journal}{Lambek Festschrift, Linguistic Analysis
  36(1--4)}}, pp. \bibinfo{pages}{345--384}.

\bibitemdeclare{article}{correia2020density}
\bibitem{correia2020density}
\bibinfo{author}{Adriana~D. \surnamestart Correia\surnameend},
  \bibinfo{author}{Michael \surnamestart Moortgat\surnameend} \&
  \bibinfo{author}{Henk~T.C. \surnamestart Stoof\surnameend}
  (\bibinfo{year}{2020}): \emph{\bibinfo{title}{Density matrices with metric
  for derivational ambiguity}}.
\newblock {\sl \bibinfo{journal}{Journal of Applied Logics}}
\bibinfo{volume}{7}(\bibinfo{number}{5}).

\bibitemdeclare{article}{greco2019vector}
\bibitem{greco2019vector}
\bibinfo{author}{Giuseppe \surnamestart Greco\surnameend}, \bibinfo{author}{Fei
  \surnamestart Liang\surnameend}, \bibinfo{author}{Michael \surnamestart
  Moortgat\surnameend} \& \bibinfo{author}{Alessandra \surnamestart
  Palmigiano\surnameend} (\bibinfo{year}{2020}): \emph{\bibinfo{title}{Vector
  spaces as Kripke frames}}.
\newblock {\sl \bibinfo{journal}{Journal of Applied Logics}}
\bibinfo{volume}{7}(\bibinfo{number}{5}).

\bibitemdeclare{book}{griffiths2018introduction}
\bibitem{griffiths2018introduction}
\bibinfo{author}{David~J. \surnamestart Griffiths\surnameend} \&
  \bibinfo{author}{Darrell~F. \surnamestart Schroeter\surnameend}
  (\bibinfo{year}{2018}): \emph{\bibinfo{title}{Introduction to quantum
  mechanics}}.
\newblock \bibinfo{publisher}{Cambridge University Press},
\doi{10.1017/9781316995433}.

\bibitemdeclare{incollection}{KurtoMM}
\bibitem{KurtoMM}
\bibinfo{author}{N.~\surnamestart Kurtonina\surnameend} \&
  \bibinfo{author}{M.~\surnamestart Moortgat\surnameend}
  (\bibinfo{year}{1997}): \emph{\bibinfo{title}{Structural Control}}.
\newblock In \bibinfo{editor}{P.~\surnamestart Blackburn\surnameend} \&
  \bibinfo{editor}{M.~\surnamestart de~Rijke\surnameend}, editors: {\sl
  \bibinfo{booktitle}{Specifying Syntactic Structures}},
  \bibinfo{publisher}{CSLI}, \bibinfo{address}{Stanford}, pp.
  \bibinfo{pages}{75--113}.

\bibitemdeclare{article}{lambek1958mathematics}
\bibitem{lambek1958mathematics}
\bibinfo{author}{Joachim \surnamestart Lambek\surnameend}
  (\bibinfo{year}{1958}): \emph{\bibinfo{title}{The mathematics of sentence
  structure}}.
\newblock {\sl \bibinfo{journal}{The American Mathematical Monthly}}
  \bibinfo{volume}{65}(\bibinfo{number}{3}),
  pp. \bibinfo{pages}{154--170},
\doi{10.1080/00029890.1958.11989160}.

\bibitemdeclare{incollection}{Lam61}
\bibitem{Lam61}
\bibinfo{author}{Joachim \surnamestart Lambek\surnameend}
  (\bibinfo{year}{1961}): \emph{\bibinfo{title}{On the calculus of syntactic
  types}}.
\newblock In \bibinfo{editor}{Roman \surnamestart Jakobson\surnameend}, editor:
  {\sl \bibinfo{booktitle}{Structure of Language and its Mathematical
  Aspects}}, {\sl \bibinfo{series}{Proceedings of Symposia in Applied
  Mathematics}} \bibinfo{volume}{XII}, \bibinfo{publisher}{American
  Mathematical Society}, pp. \bibinfo{pages}{166--178},
\doi{10.1090/psapm/012/9972}.

\bibitemdeclare{article}{MontagueUG}
\bibitem{MontagueUG}
\bibinfo{author}{Richard \surnamestart Montague\surnameend}
  (\bibinfo{year}{1970}): \emph{\bibinfo{title}{Universal grammar}}.
\newblock {\sl \bibinfo{journal}{Theoria}}
  \bibinfo{volume}{36}(\bibinfo{number}{3}),
  pp. \bibinfo{pages}{373--398},
\doi{10.1111/j.1755-2567.1970.tb00434.x}.

\bibitemdeclare{article}{mmjolli}
\bibitem{mmjolli}
\bibinfo{author}{Michael \surnamestart Moortgat\surnameend}
  (\bibinfo{year}{1996}): \emph{\bibinfo{title}{Multimodal Linguistic
  Inference}}.
\newblock {\sl \bibinfo{journal}{Journal of Logic, Language and Information}}
  \bibinfo{volume}{5}(\bibinfo{number}{3/4}),
  pp. \bibinfo{pages}{349--385},
\doi{10.1007/BF00159344}.

\bibitemdeclare{article}{moortgat2019frobenius}
\bibitem{moortgat2019frobenius}
\bibinfo{author}{Michael \surnamestart Moortgat\surnameend},
  \bibinfo{author}{Mehrnoosh \surnamestart Sadrzadeh\surnameend} \&
  \bibinfo{author}{Gijs \surnamestart Wijnholds\surnameend}
  (\bibinfo{year}{2020}): \emph{\bibinfo{title}{A {F}robenius Algebraic
  Analysis for Parasitic Gaps}}.
\newblock {\sl \bibinfo{journal}{Journal of Applied Logics}}
  \bibinfo{volume}{7}(\bibinfo{number}{5}), pp. \bibinfo{pages}{823--852}.

\bibitemdeclare{article}{moortgat2017lexical}
\bibitem{moortgat2017lexical}
\bibinfo{author}{Michael \surnamestart Moortgat\surnameend} \&
  \bibinfo{author}{Gijs \surnamestart Wijnholds\surnameend}
  (\bibinfo{year}{2017}): \emph{\bibinfo{title}{Lexical and derivational
  meaning in vector-based models of relativisation}}.
\newblock {\sl \bibinfo{journal}{arXiv:\href{https://arxiv.org/abs/1711.11513}{1711.11513} }}.

\bibitemdeclare{book}{nielsen2002quantum}
\bibitem{nielsen2002quantum}
\bibinfo{author}{Michael~A \surnamestart Nielsen\surnameend} \&
  \bibinfo{author}{Isaac \surnamestart Chuang\surnameend}
  (\bibinfo{year}{2002}): \emph{\bibinfo{title}{Quantum computation and quantum
  information}}.
\newblock \bibinfo{publisher}{Cambridge University Press}.

\bibitemdeclare{article}{piedeleuopen}
\bibitem{piedeleuopen}
\bibinfo{author}{Robin \surnamestart Piedeleu\surnameend},
  \bibinfo{author}{Dimitri \surnamestart Kartsaklis\surnameend},
  \bibinfo{author}{Bob \surnamestart Coecke\surnameend} \&
  \bibinfo{author}{Mehrnoosh \surnamestart Sadrzadeh\surnameend}
  (\bibinfo{year}{2015}): \emph{\bibinfo{title}{Open System Categorical Quantum
  Semantics in Natural Language Processing}}.
\newblock {\sl \bibinfo{journal}{6th International Conference on Algebra and
  Coalgebra in Computer Science (CALCO’15). Ed: Larry Moss and Paweł
  Sobociński}}, p. \bibinfo{pages}{267–286}, \doi{10.4230/LIPIcs.CALCO.2015.267}.


\bibitemdeclare{article}{DBLP:journals/amai/SadrzadehKB18}
\bibitem{DBLP:journals/amai/SadrzadehKB18}
\bibinfo{author}{Mehrnoosh \surnamestart Sadrzadeh\surnameend},
  \bibinfo{author}{Dimitri \surnamestart Kartsaklis\surnameend} \&
  \bibinfo{author}{Esma \surnamestart Balkir\surnameend}
  (\bibinfo{year}{2018}): \emph{\bibinfo{title}{Sentence entailment in
  compositional distributional semantics}}.
\newblock {\sl \bibinfo{journal}{Ann. Math. Artif. Intell.}}
  \bibinfo{volume}{82}(\bibinfo{number}{4}),
  pp. \bibinfo{pages}{189--218},
\doi{10.1162/coli.2006.32.3.379}.

\bibitemdeclare{inproceedings}{wansing1992}
\bibitem{wansing1992}
\bibinfo{author}{Heinrich \surnamestart Wansing\surnameend}
  (\bibinfo{year}{1992}): \emph{\bibinfo{title}{Formulas-as-types for a
  hierarchy of sublogics of intuitionistic propositional logic}}.
\newblock In \bibinfo{editor}{David \surnamestart Pearce\surnameend} \&
  \bibinfo{editor}{Heinrich \surnamestart Wansing\surnameend}, editors: {\sl
  \bibinfo{booktitle}{Nonclassical Logics and Information Processing}},
  \bibinfo{publisher}{Springer Berlin Heidelberg}, pp.
  \bibinfo{pages}{125--145},
\doi{10.1007/BFb0031928}.

\end{thebibliography}

\appendix

\section{Complete proof trees for Dutch relativization clauses}\label{fulltrees}

\subsection{Subject Relativization}

\begin{center}
\begin{equation*}\scalebox{0.7}{
\hspace{-2.2cm}\infer[\bo \bs E \bc^{}]{\mbox{man}\cdot_{}(\mbox{die}\cdot_{}((\mbox{de}\cdot_{}\mbox{hond})\cdot_{}\mbox{bijt})) \vdash \textcolor{red}{(y_{0} \triangleright (z_{0} \triangleleft \lambda x_{1}. ({}^{\vee}  {}^{\cup} x_{1} \triangleright ((x_{2} \triangleleft \  y_{2}) \triangleright z_{2}))))} : n}{
      \infer[\ell]{\textcolor{red}{y_{0}} : n}{\mbox{man}}
   & 
      \infer[\bo / E \bc^{}]{\mbox{die}\cdot_{}((\mbox{de}\cdot_{}\mbox{hond})\cdot_{}\mbox{bijt}) \vdash \textcolor{red}{(z_{0} \triangleleft \lambda x_{1}. ({}^{\vee}  {}^{\cup} x_{1} \triangleright ((x_{2} \triangleleft \  y_{2}) \triangleright z_{2})))} : n \bs_{}n}{
         \infer[\ell]{\textcolor{red}{z_{0}} : (n \bs_{}n) /_{}(\diamondsuit_{}\Box_{}np \bs_{}s)}{\mbox{die}}
      & 
         \infer[\bo \bs I \bc^{1}]{(\mbox{de}\cdot_{}\mbox{hond})\cdot_{}\mbox{bijt} \vdash \textcolor{red}{\lambda x_{1}. ({}^{\vee}  {}^{\cup} x_{1} \triangleright ((x_{2} \triangleleft \  y_{2}) \triangleright z_{2}))} : \diamondsuit_{}\Box_{}np \bs_{}s}{
            \infer[\bo \diamondsuit E \bc^{2}]{\makebox[.85em]{\textvisiblespace}\cdot_{}((\mbox{de}\cdot_{}\mbox{hond})\cdot_{}\mbox{bijt}) \vdash \textcolor{red}{( {}^{\vee}  {}^{\cup} x_{1} \triangleright ((x_{2} \triangleleft \  y_{2}) \triangleright z_{2}))} : s}{
                \bo \makebox[.85em]{\textvisiblespace} \vdash \textcolor{red}{x_{1}} : \diamondsuit_{}\Box_{}np \bc^{1} 
            & 
               \infer[\bo \bs E \bc^{}]{\langle \makebox[.85em]{\textvisiblespace}\rangle^{}\cdot_{}((\mbox{de}\cdot_{}\mbox{hond})\cdot_{}\mbox{bijt}) \vdash \textcolor{red}{({}^{\vee} z_{1} \triangleright ((x_{2} \triangleleft \  y_{2}) \triangleright z_{2}))} : s}{
                  \infer[\bo \Box E \bc^{}]{\langle \makebox[.85em]{\textvisiblespace}\rangle^{} \vdash \textcolor{red}{ {}^{\vee} z_{1}} : np}{
                      \bo \makebox[.85em]{\textvisiblespace} \vdash \textcolor{red}{z_{1}} : \Box_{}np \bc^{2} 
                  }
               & 
                  \infer[\bo \bs E \bc^{}]{(\mbox{de}\cdot_{}\mbox{hond})\cdot_{}\mbox{bijt} \vdash \textcolor{red}{((x_{2} \triangleleft \  y_{2}) \triangleright z_{2})} : np \bs_{}s}{
                     \infer[\bo / E \bc^{}]{\mbox{de}\cdot_{}\mbox{hond} \vdash \textcolor{red}{(x_{2} \triangleleft \  y_{2})} : np}{
                        \infer[\ell]{\textcolor{red}{x_{2}} : np /_{}n}{\mbox{de}}
                     & 
                        \infer[\ell]{\textcolor{red}{y_{2}} : n}{\mbox{hond}}
                     }
                  & 
                     \infer[\ell]{\textcolor{red}{z_{2}} : np \bs_{}(np \bs_{}s)}{\mbox{bijt}}
                  }
               }
            }
         }
      }
   }
   }
\end{equation*}
\end{center}

\subsection{Object Relativization}

\begin{center}
\begin{equation*}
\scalebox{0.7}{
\hspace{-2.2cm}\infer[\bo \bs E \bc^{}]{\mbox{man}\cdot_{}(\mbox{die}\cdot_{}((\mbox{de}\cdot_{}\mbox{hond})\cdot_{}\mbox{bijt})) \vdash \textcolor{red}{( y_{0} \triangleright (z_{0} \triangleleft \lambda x_{1}.((x_{2}   \triangleleft y_{2}) \triangleright (   {}^{\vee}  {}^{\cup} x_{1} \triangleright z_{2}))))} : n}{
      \infer[\ell]{\textcolor{red}{y_{0}} : n}{\mbox{man}}
   & 
      \infer[\bo / E \bc^{}]{\mbox{die}\cdot_{}((\mbox{de}\cdot_{}\mbox{hond})\cdot_{}\mbox{bijt}) \vdash \textcolor{red}{(z_{0} \triangleleft \lambda x_{1}.((x_{2}   \triangleleft y_{2}) \triangleright (   {}^{\vee}  {}^{\cup} x_{1} \triangleright z_{2})))} : n \bs_{}n}{
         \infer[\ell]{\textcolor{red}{z_{0}} : (n \bs_{}n) /_{}(\diamondsuit_{}\Box_{}np \bs_{}s)}{\mbox{die}}
      & 
         \infer[\bo \bs I \bc^{1}]{(\mbox{de}\cdot_{}\mbox{hond})\cdot_{}\mbox{bijt} \vdash \textcolor{red}{\lambda x_{1}.((x_{2}   \triangleleft y_{2}) \triangleright (   {}^{\vee}  {}^{\cup} x_{1} \triangleright z_{2}))} : \diamondsuit_{}\Box_{}np \bs_{}s}{
            \infer[\bo \diamondsuit E \bc^{2}]{\makebox[.85em]{\textvisiblespace}\cdot_{}((\mbox{de}\cdot_{}\mbox{hond})\cdot_{}\mbox{bijt}) \vdash \textcolor{red}{((x_{2}   \triangleleft y_{2}) \triangleright (   {}^{\vee}  {}^{\cup} x_{1} \triangleright z_{2}))} : s}{
                \bo \makebox[.85em]{\textvisiblespace} \vdash \textcolor{red}{x_{1}} : \diamondsuit_{}\Box_{}np \bc^{1} 
            & 
               \infer[\bo Comm_\lozenge \bc^{}]{\langle \makebox[.85em]{\textvisiblespace}\rangle^{}\cdot_{}((\mbox{de}\cdot_{}\mbox{hond})\cdot_{}\mbox{bijt}) \vdash \textcolor{red}{((x_{2}   \triangleleft y_{2}) \triangleright (   {}^{\vee} z_{1} \triangleright z_{2}))} : s}{
                  \infer[\bo \bs E \bc^{}]{(\mbox{de}\cdot_{}\mbox{hond})\cdot_{}(\langle \makebox[.85em]{\textvisiblespace}\rangle^{}\cdot_{}\mbox{bijt}) \vdash \textcolor{red}{((x_{2}   \triangleleft y_{2}) \triangleright (   {}^{\vee} z_{1} \triangleright z_{2}))} : s}{
                     \infer[\bo / E \bc^{}]{\mbox{de}\cdot_{}\mbox{hond} \vdash \textcolor{red}{(x_{2}   \triangleleft y_{2})} : np}{
                        \infer[\ell]{\textcolor{red}{x_{2}} : np /_{}n}{\mbox{de}}
                     & 
                        \infer[\ell]{\textcolor{red}{y_{2}} : n}{\mbox{hond}}
                     }
                  & 
                     \infer[\bo \bs E \bc^{}]{\langle \makebox[.85em]{\textvisiblespace}\rangle^{}\cdot_{}\mbox{bijt} \vdash \textcolor{red}{(   {}^{\vee} z_{1} \triangleright z_{2})} : np \bs_{}s}{
                        \infer[\bo \Box E \bc^{}]{\langle \makebox[.85em]{\textvisiblespace}\rangle^{} \vdash \textcolor{red}{ {}^{\vee} z_{1}} : np}{
                            \bo \makebox[.85em]{\textvisiblespace} \vdash \textcolor{red}{z_{1}} : \Box_{}np \bc^{2} 
                        }
                     & 
                        \infer[\ell]{\textcolor{red}{z_{2}} : np \bs_{}(np \bs_{}s)}{\mbox{bijt}}
                     }
                  }
               }
            }
         }
      }
   }
   }
\end{equation*}
\end{center}

\subsection{Formal semantics of relative pronouns}

To obtain the usual `formal semantics' terms, one substitutes for the parameter $z_0$ the lexical program for the word `die':

\[\textsc{die} = \lambda x\lambda y\lambda z.((y\ z) \wedge (x\ {}^{\cap}{}^{\wedge} z))\]
which then, after $\beta$ conversion and cap-cup and wedge-vee cancellation, reduces to

\[\textcolor{red}{\lambda z_{}.((\textsc{man} \  z_{}) \wedge ((\textsc{bijt} \  (\textsc{de} \  \textsc{hond})) \  z_{}))}\quad\textrm{(subject reading)}\]
\[\textcolor{red}{\lambda z_{}.((\textsc{man} \  z_{}) \wedge ((\textsc{bijt} \  z_{}) \  (\textsc{de} \  \textsc{hond})))}\quad\textrm{(object reading)}\]

\section{Interpretation of extended $[\textit{xleft}]^n$ rule}\label{xleftinterp}

To arrive at the interpretation of the \textit{xleft} rule, we compose the interpretations of the rules that it abreviates, explicit on the left part of \ref{xleftfirst}. Additionally to the interpretations of $E_{\square}$, $E_{\lozenge}$ and $I_{\backslash}$, we only need to provide the interpretation for \textit{Ass}$_{\lozenge}$ and \textit{Comm}$_{\lozenge}$. Strucutral rules do not affect systematically the programme encoded by the associated lambda term. Howver, in this paper we go beyond the "bag of words" view and introduce a specification in the lambda term that results from the \textit{Comm}$_{\lozenge}$ rule:

\begin{equation}
\infer[Comm_\lozenge]{\nd{\Gamma[\langle \Delta_1 \rangle\cdott(\Delta_2\cdott \Delta_3)]}{
^ct:B}}{\nd{\Gamma[\Delta_2\cdott(\langle \Delta_1 \rangle\cdott \Delta_3)]}{ t:B} }
\end{equation}

The interpretation of \textit{Comm}$_{\lozenge}$ is as follows:

$$\llbracket (^c t)^B \rrbracket_g = \llbracket t^{B} \rrbracket_{g^S} \otimes S_+ \left(\llbracket t^{B} \rrbracket_{g^{\mathfrak{S}}} \right) (S_+)^\dagger,$$ with $S_+$ the raising operator in the interpreting space, according to the discussion in sec. \ref{structural}.  If it is applied $n$ times successively , it is takes the form and respective interpretation 

$$\left\llbracket \left({}^{c^n} t\right)^B \right\rrbracket_g = \llbracket t^{B} \rrbracket_{g^S} \otimes \left(S_+\right)^n \left(\llbracket t^{B} \rrbracket_{g^{\mathfrak{S}}} \right) \left((S_+)^\dagger\right)^n.$$ This extends naturally to the case when the \textit{Comm}$_{\lozenge}$ rule is never applied, in which case $n=0$, where we have that $\left(S_+\right)^0=I.$

In what follows we take the necessary steps to arrive at the interpretation of term $\lambda^{l} x.{}^{c^n}t[{}^{\cup} x/z]$ in spin space. First, we interpret the application of \textit{Comm}$_{\lozenge}$:

$$\left\llbracket \left({}^{c^n}t\right)^B[{}^{\cup} x/z] \right\rrbracket_{g^{\mathfrak{S}}} = \left(S_+\right)^n \llbracket \left(t[{}^{\cup} x/z]\right)^B \rrbracket_{g^{\mathfrak{S}}} \left((S_+)^\dagger\right)^n $$

Then we expand on the interpretation of $E_\lozenge$:

$$\llbracket \left(t[{}^{\cup} x/z]\right)^B \rrbracket_{g^{\mathfrak{S}}} = \llbracket t^B \rrbracket_{g^{\mathfrak{S}}_{z,^\cup x}},$$ which means that 

\begin{equation}\label{xinter}
\llbracket (^\cup x)^{\square A} \rrbracket_{g^{\mathfrak{S}}} = T'^0_\lozenge \left(\llbracket x^{\lozenge \square A} \right)  \rrbracket_{g^{\mathfrak{S}}}
\end{equation}

will replace $\llbracket z^{\square A} \rrbracket$ inside of $t$, appearing here already as the result of the application of $E_\square$:

$$\llbracket (^\vee z)^{A} \rrbracket_{g^{\mathfrak{S}}} = T'^0_\square \left(\llbracket z^{\square A}  \rrbracket_{g^{\mathfrak{S}}} \right) .$$

Finally, abstracting over variable $x$ is interpreted as 

$$\llbracket \lambda^{l} x.{}^{c^n}t[{}^{\cup} x/z] \rrbracket_{g^{\mathfrak{S}}} =  \llbracket {}^{c^n}t[{}^{\cup} x/z] \rrbracket_{g^{\mathfrak{S}}_{x,I}},$$
 such that the only instance of $x$ has its interpretation subsituted by the indentity, namely in eq.(\ref{xinter}). Putting all these elements together and normalizing, we arrive at the interpretation in eq.(\ref{xleftprev}).

\section{Concrete interpretation of relative clauses}\label{fullinterpretation}

The derivations in \ref{types} have a final term that depends on the variables $y_0$, $z_0$, $x_2$, $y_2$ $z_2$ and $x_1$. The latter is a \textit{bound} variable (as well as the intermediate variable $x$), due to the lambda abstraction, and the former are \textit{free} variables. Bound variables can be substituted by any free variable during the derivation, via beta reduction, and will take the value of that variable, contrasting with free variables that will be substituted by constants, and interpreted accordingly. An assignment function $g$ assigns bound variables to a later-to-be-defined constant, and assigns free variables to specific constants, here our words. In our assignment, taken as an example, the assignment function gives $g(y_0)=man'$ but $g(x_1)$ remain in this form, until $x_1$ is substituted by a free variable. Alternatively we can represent the free variables as bound variables using a lambda abstraction, applied on a constant: $\lambda y_0. y_0 (man') \rightarrow man'$.

Looking at the interpretation of any variable stated in the interpretation of the axiom rule in eq.(\ref{axiom}) and comparing with the interpretation of the constants in eqs.(\ref{man}) to (\ref{bijt}), we note that both represent the density matrix entries in a symbolic form, where we can apply directly operations like trace and matrix multiplication in the spatial components, or spin operators in the spin components. This permits that, when we perform these calculation step by step using each rule, we can perform them directly on the symbolic representations of interpretations of constants, in eqs.(\ref{man}) to (\ref{bijt}), as well as of variables that naturaly take the same form as states in eq.(\ref{axiom}), since it can potentially take the value of \textit{any} other constant.

Therefore, one can impose an assignment that will interpret our particular Dutch relative clause "man die de hond bijt" $g$ that instantiates the free variables like so:

\begin{align}
&\llbracket (x_2 \triangleleft y_2) \rrbracket_g = \llbracket \text{de\_hond'}^{np} \rrbracket_I, \\
&\llbracket (z_2) \rrbracket_g = \llbracket \text{bijt'}^{np\backslash np\backslash s} \rrbracket_I, \\
&\llbracket z_0 \rrbracket_g = \llbracket \text{die'}^{(n\backslash)\slash (np \backslash s)} \rrbracket_I, \\
&\llbracket y_0 \rrbracket_g=\llbracket \text{man'}^n \rrbracket_I 
\end{align} and instantiates the bound variable $x$ according to eq.(\ref{axiom}).

Substituting these directly in the derivations, we can, step by step, arrive at the final different readings. In what follows we give a full breakdown of these steps, splitting between spatial and spin components, and between subject and object relativization.

\subsection{Interpretations in $\lceil . \rceil$:}\label{distinter}

\subsubsection{Subject Relativization}

The interpretation of this derivation starts by making use of the interpretation of $E_\backslash$ as given in eq.(\ref{triangleright}), substituting the variables by the assigned constants as described above.

\begin{flalign}
   & \llbracket (x_2 \triangleleft y_2) \triangleright z_2 \rrbracket_{g^S} = \Tr_{\tilde{N}} \left( \llbracket (x_2 \triangleleft y_2) \rrbracket_{g^S} \cdot \llbracket  z_2 \rrbracket_{g^S} \right) \nonumber \\
    &\Tr_{\tilde{N}} \left( \sum_{jj'} {^S \textbf{H}^{jj'}} \ket{_{j}}\prescript{}{\tilde{N}}{\bra{_{j'}}} \cdot \sum_{oo', pp', qq'} {^S\textbf{B}_{o'o, p'p}^{\;\;\; \; \;\; \; \;\;qq'}} \ket{^{o'p'}_{\;\;\; \; \;\;q}} \prescript{}{\lceil N \rceil^* \otimes \lceil N \rceil^* \otimes \lceil S \rceil}{\bra{^{o p}_{\;\;\; \; q'}}} \right) \nonumber \\
   & =\sum_{jj', pp', qq'} {^S \textbf{H}^{jj'}} {^S\textbf{B}_{j'j, p'p}^{\;\;\; \; \;\; \; \;\;qq'}} \ket{^{p'}_{\;\;\; \; \;\;q}} \prescript{}{ \lceil N \rceil^* \otimes \lceil S \rceil}{\bra{^{ p}_{\;\;\; \; q'}}}
\end{flalign}

Then we use again eq.(\ref{triangleright}) and the interpret the variable $x$ using axiom rule as in eq.(\ref{axiom}). 

\begin{align} \label{order1}
& \llbracket x \triangleright ((x_2 \triangleleft y_2) \triangleright z_2) \rrbracket_{g^S} =\Tr_{\tilde{N}} \left( \llbracket x \rrbracket_{g^S} \cdot \llbracket (x_2 \triangleleft y_2) \triangleright z_2 \rrbracket_{g^S}  \right) \nonumber \\
 &  = \Tr_{\tilde{N}} \left( \sum_{ii'} {^S \textbf{X}^{ii'}} \ket{_{i}}\prescript{}{\tilde{N}}{\bra{_{i'}}} \cdot \sum_{jj', pp', qq'} {^S\textbf{H}^{jj'}} {^S\textbf{B}_{j'j, p'p}^{\;\;\; \; \;\; \; \;\;qq'}} \ket{^{p'}_{\;\;\; \; \;\;q}} \prescript{}{\lceil N \rceil^* \otimes  \ \lceil S \rceil}{\bra{^{ p}_{\;\;\; \; q'}}} \right) \nonumber \\
   & = \sum_{ii',jj', qq'} {^S \textbf{X}^{ii'}} {^S\textbf{H}^{jj'}} {^S\textbf{B}_{j'j, i'i}^{\;\;\; \; \;\; \; \;\;qq'}} \ket{^{}_{q}} \prescript{}{ \lceil S \rceil}{\bra{^{}_{q'}}}
\end{align}

To use the \textit{xleft} rule, we first interpret the previous term in the assignment $g_{x,ll'}^S$, as described in Def.\ref{altg}. recalculating the previous interpretation using the basis of its interpretation space instead of eq.(\ref{axiom}).

\begin{align}
     & \llbracket x \triangleright ((x_2 \triangleleft y_2) \triangleright z_2) \rrbracket_{g_{x,ll'}^S}  = \Tr_{\tilde{N}} \left( \ket{_{l}} \prescript{}{\lceil N \rceil}{\bra{_{l'}}}  \cdot \llbracket (x_2 \triangleleft y_2) \triangleright z_2 \rrbracket_{g^S}  \right) \nonumber \\
     & = \sum_{jj', qq'}  {^S\textbf{H}^{jj'}} {^S\textbf{B}_{j'j, l'l}^{\;\;\; \; \;\; \; \;\;qq'}} \ket{^{}_{ q}} \prescript{}{ \lceil S \rceil}{\bra{^{}_{ q'}}}
\end{align}

We simplify the spatial interpretation of \textit{xleft} as given in eq.(\ref{xleftprev}), using that $x$ and $y$ are interpreted both interpreted in $\lceil A \rceil$, since $\lceil \lozenge \square A \rceil=  \lceil A \rceil$:

\begin{align}\label{xleftx}
 & \llbracket (\lambda^{l} x.{}^{c^0}t[{}^\vee {}^{\cup} x/y])^{\fdia\gbox A\bs B} \rrbracket_{g^S} = \nonumber \\
 & = \sum_{ll'} \ket{^{l'}_{}} \prescript{}{\lceil A \rceil^*}{\bra{^{l}}} \otimes \left[\Tr_{\lceil A \rceil} \left( \left\llbracket  x^{\lozenge \square A} \right\rrbracket_{g^{S}} \cdot \sum_{kk'}   \ket{^{k'}_{}} \prescript{}{\lceil A \rceil^*}{\bra{^{k}}} \otimes \llbracket t^B \rrbracket_{g^{S}_{y,kk'}} \right) \right]_{g^{S}_{x,ll'}} \nonumber \\ 
    & = \sum_{ll'} \ket{^{l'}_{}} \prescript{}{\lceil A \rceil^*}{\bra{^{l}}} \otimes \Tr_{\lceil A \rceil} \left( \ket{_{l}} \prescript{}{\lceil A \rceil}{\bra{_{l'}}} \cdot \sum_{kk'}   \ket{^{k'}_{}} \prescript{}{\lceil A \rceil^*}{\bra{^{k}}} \otimes \llbracket t^B \rrbracket_{g^{S}_{y,kk'}} \right)
    \nonumber \\
    & = \sum_{ll'} \ket{^{l'}_{}} \prescript{}{\lceil A \rceil^*}{\bra{^{l}}} \otimes  \llbracket t^B \rrbracket_{g^{S}_{y,ll'}}.
\end{align} Using this simplified form, we see that multiplying with the dual basis of the space that interprets both $x$ and $x_1$ results in an expression that will take any value of a variable of that type, precisely the goal of the lambda abstraction.

\begin{align}
&\llbracket \lambda^{l} x_{1}.{}^{c^0}t({}^{\vee}  {}^{\cup} x_{1}\triangleright ((x_{2} \triangleleft  y_{2})\triangleright z_{2})) \rrbracket_{g^S}=  \sum_{ll'} \ket{^{l'}_{}} \prescript{}{\lceil N \rceil^*}{\bra{^{l}}} \otimes  \llbracket x \triangleright ((x_2 \triangleleft y_2) \triangleright z_2) \rrbracket_{g_{x,ll'}^S} \nonumber \\
& = \sum_{ll'} \ket{^{l'}_{}} \prescript{}{\lceil N \rceil^*}{\bra{^{l}}} \otimes \sum_{jj', qq'}  {^S\textbf{H}^{jj'}} {^S\textbf{B}_{j'j, l'l}^{\;\;\; \; \;\; \; \;\;qq'}} \ket{^{}_{q}} \prescript{}{ \lceil S \rceil}{\bra{^{}_{ q'}}} \nonumber \\
& = \sum_{ll', jj', qq'} {^S\textbf{H}^{jj'}} {^S\textbf{B}_{j'j, l'l}^{\;\;\; \; \;\; \; \;\;qq'}} \ket{^{l'}_{ \; \; q}} \prescript{}{ \lceil N \rceil^* \otimes \lceil S \rceil}{\bra{^{l}_{ \; \; q'}}}
\end{align}

To finalize, the next two steps consist in the application of the interpretations of $E_\slash$   in eq.(\ref{triangleleft}) and $E_\backslash$ (eq.\ref{triangleright}), respectively, resulting in the spatial part of eq. \ref{subrel}.

\begin{align}
&\llbracket z_{0} \triangleleft \lambda^{l} x_{1}.{}^{c^0}t({}^{\vee}  {}^{\cup} x_{1}\triangleright ((x_{2} \triangleleft  y_{2})\triangleright z_{2})) \rrbracket_{g^S}  = \Tr_{\tilde{S}} \left( \Tr_{\tilde{N}} \left( \llbracket z_{0} \rrbracket_{g^S} . \llbracket \lambda^{l} x_{1}.{}^{c^0}t({}^{\vee}  {}^{\cup} x_{1}\triangleright ((x_{2} \triangleleft  y_{2})\triangleright z_{2})) \rrbracket_{g^S} \right) \right) \nonumber \\
& =  \Tr_{\tilde{S}} \left( \Tr_{\tilde{N}} \left( \sum_{kk', tt',mm', nn'} {^S\textbf{D}_{k'k\;m'm}^{\; \; \; tt'\; \; \;nn'}} \ket{^{k'\;m'}_{\; \; \; t\; \; \;n} } \prescript{}{\lceil N \rceil^* \otimes \lceil N \rceil \otimes (\lceil S \rceil^* \otimes \lceil N \rceil)}{\bra{^{k\;m}_{\; \;t'\; \;n'\;}}}  \right. \right. \nonumber \\
& \left. \left. \cdot \sum_{ll', jj', qq'} {^S\textbf{H}^{jj'}} {^S\textbf{B}_{j'j, l'l}^{\;\;\; \; \;\; \; \;\;qq'}} \ket{^{l'}_{ \; \; q}} \prescript{}{ \lceil N \rceil^* \otimes \lceil S \rceil}{\bra{^{l}_{ \; \; q'}}} \right) \right) \nonumber \\
& = \sum_{kk', tt',mm', nn', jj'} {^S\textbf{D}_{k'k\;m'm}^{\; \; \; tt'\; \; \;nn'}} {^S\textbf{H}^{jj'}} {^S\textbf{B}_{j'j, n'n}^{\;\;\; \; \;\; \; \;\;mm'}}  \ket{^{k'}_{\; \; \; t} } \prescript{}{\lceil N \rceil^* \otimes \lceil N \rceil}{\bra{^{k}_{\; \;t'\;}}}
\end{align}

\begin{align}
&\llbracket y_{0}\triangleright (z_{0} \triangleleft \lambda^{l} x_{1}.{}^{c^0}t({}^{\vee}  {}^{\cup} x_{1}\triangleright ((x_{2} \triangleleft  y_{2})\triangleright z_{2}))) \rrbracket_{g^S} = \Tr_{\tilde{N}} \left(  \llbracket y_{0} \rrbracket_{g^S} \cdot \llbracket z_{0} \triangleleft \lambda^{l} x_{1}.{}^{c^0}t({}^{\vee}  {}^{\cup} x_{1}\triangleright ((x_{2} \triangleleft  y_{2})\triangleright z_{2})) \rrbracket_{g^S}  \right) \nonumber \\
& \Tr_{\tilde{N}} \left( \sum_{rr'} {^S\textbf{M}}^{rr'} \ket{_{r}}\prescript{}{\lceil N \rceil}{\bra{_{r'}}} \cdot \sum_{kk', tt',mm', nn', jj'} {^S\textbf{D}_{k'k\;m'm}^{\; \; \; tt'\; \; \;nn'}} {^S\textbf{H}^{jj'}} {^S\textbf{B}_{j'j, n'n}^{\;\;\; \; \;\; \; \;\;mm'}}  \ket{^{k'}_{\; \; \; t}} \prescript{}{\lceil N \rceil^* \otimes \lceil N \rceil}{\bra{^{k}_{\; \;t'\;}}} \right) \nonumber\\
& =  \sum_{rr', tt',mm', nn', jj'} {^S\textbf{M}}^{rr'} {^S\textbf{D}_{r'r\;m'm}^{\; \; \; tt'\; \; \;nn'}} {^S\textbf{H}^{jj'}} {^S\textbf{B}_{j'j, n'n}^{\;\;\; \; \;\; \; \;\;mm'}} \ket{^{}_{t}} \prescript{}{ \lceil N \rceil}{\bra{^{}_{t'}}}\\
&= \llbracket \text{man\_die\_de\_hond\_bijt'} \rrbracket^1_{I^S}
\end{align}

\subsubsection{Object relativization}

This derivation is very similar to the previous, except that on the first application of $E_\backslash$ the bound variable $x$ is introduced as the argument of $z_2$, and only on the next application of the rule is $(x_{2} \triangleleft y_{2})$ taken as an argument.

\begin{align}
 &\llbracket x \triangleright z_2 \rrbracket_{g^S} =\Tr_{\tilde{N}} \left( \llbracket x \rrbracket_{g^S} \cdot \llbracket z_2 \rrbracket_{g^S}  \right) \nonumber \\
   & = \Tr_{\tilde{N}} \left( \sum_{ii'} {^S \textbf{X}^{ii'}} \ket{_{i}}\prescript{}{\tilde{N}}{\bra{_{i'}}} \cdot\sum_{oo', pp', qq'} {^S\textbf{B}_{o'o, p'p}^{\;\;\; \; \;\; \; \;\;qq'}} \ket{^{o'p'}_{\;\;\; \; \;\;q}} \prescript{}{\lceil N \rceil^* \otimes \lceil N \rceil^* \otimes \lceil S \rceil}{\bra{^{o p}_{\;\;\; \; q'}}} \right) \nonumber \\
   & = \sum_{ii',pp', qq'} {^S \textbf{X}^{ii'}} {^S\textbf{B}_{i'i, p'p}^{\;\;\; \; \;\; \; \;\;qq'}} \ket{^{p'}_{\;\;\; \; \;\;q}} \prescript{}{\lceil N \rceil^* \otimes \lceil S \rceil}{\bra{^{p}_{\;\;\; \; q'}}}
\end{align}

\begin{align}
 & \llbracket (x_{2} \triangleleft y_{2})\triangleright(x \triangleright z_{2}) \rrbracket_{g^S}= \Tr_{\tilde{N}} \left( \llbracket (x_{2} \triangleleft y_{2}) \rrbracket_{g^S} \cdot \llbracket x \triangleright z_{2} \rrbracket_{g^S}  \right) = \Tr_{\tilde{N}} \left( \llbracket (x_{2} \triangleleft y_{2}) \rrbracket_{g^S} \cdot \Tr_{\tilde{N}} \left( \llbracket x \rrbracket_{g^S} \cdot \llbracket z_2 \rrbracket_{g^S}  \right)  \right)  \nonumber \\
 &= \Tr_{\tilde{N}} \left( \sum_{jj'} {^S \textbf{H}^{jj'}} \ket{_{j}}\prescript{}{\tilde{N}}{\bra{_{j'}}} \cdot \sum_{ii',pp', qq'} {^S \textbf{X}^{ii'}} {^S\textbf{B}_{i'i, p'p}^{\;\;\; \; \;\; \; \;\;qq'}} \ket{^{p'}_{\;\;\; \; \;\;q}} \prescript{}{ \lceil N \rceil^* \otimes \lceil S \rceil}{\bra{^{p}_{\;\;\; \; q'}}} \right) \nonumber \\
 &= \sum_{jj', ii', qq'} {^S \textbf{H}^{jj'}} {^S \textbf{X}^{ii'}} {^S\textbf{B}_{i'i, j'j}^{\;\;\; \; \;\; \; \;\;qq'}} \ket{^{}_{q}} \prescript{}{\lceil S \rceil}{\bra{^{}_{ q'}}}
 \end{align}
 
Note at this point that, due to changing the ordering of contraction, when compared with the subject relativization reading, the matrix indices are contracted differently from eq.\ref{order1}. We see now what the role of the hypotheses $x$ is: to block $(x_2 \triangleright y_2)$ from contracting inevitably as the first argument of $z_2$. Now that the contraction is in line with what we want for an object relativization reading, we will extract variable $x$ via \textit{xleft}. To do that, we first reinterpret the previous term using the assignment $g_{x,ll'}^S$. To substitute the interpretation of $x$ by that of its basis elements we need to go further into de proof, when compared with the subject relativization reading.

\begin{align}
   &\llbracket (x_{2} \triangleleft y_{2})\triangleright(x \triangleright z_{2}) \rrbracket_{g_{x,ll'}^S} = \Tr_{\tilde{N}} \left( \llbracket (x_{2} \triangleleft y_{2}) \rrbracket_{g^S} \cdot \Tr_{\tilde{N}} \left(\ket{_{l}} \prescript{}{\lceil N \rceil}{\bra{_{l'}}}  \cdot \llbracket z_2 \rrbracket_{g^S}  \right)  \right) \nonumber \\
   &= \Tr_{\tilde{N}} \left( \sum_{jj'} {^S \textbf{H}^{jj'}} \ket{_{j}}\prescript{}{\tilde{N}}{\bra{_{j'}}} \cdot \sum_{pp', qq'}  {^S\textbf{B}_{l'l, p'p}^{\;\;\; \; \;\; \; \;\;qq'}} \ket{^{p'}_{\;\;\; \; \;\;q}} \prescript{}{\lceil N \rceil^* \otimes \lceil S \rceil}{\bra{^{p}_{\;\;\; \; q'}}} \right) \nonumber \\
   &= \sum_{jj',qq'} {^S \textbf{H}^{jj'}} {^S\textbf{B}_{l'l, j'j}^{\;\;\; \; \;\; \; \;\;qq'}} \ket{^{}_{q}} \prescript{}{ \lceil S \rceil}{\bra{^{}_{q'}}}
\end{align}

The following steps are as before, with the final result referring to eq.\ref{objrel}.

\begin{align}
&\llbracket \lambda^{l} x_{1}.{}^{c^1}t((x_{2} \triangleleft y_{2})\triangleright({}^{\vee}{}^{\cup}x_{1}\triangleright z_{2})) \rrbracket_{g^S} =  \sum_{ll'} \ket{^{l'}_{}} \prescript{}{\lceil N \rceil^*}{\bra{^{l}}} \otimes  \llbracket ((x_{2} \triangleleft y_{2})\triangleright(x \triangleright z_{2})) \rrbracket_{g_{x,ll'}^S}  \nonumber \\
& = \sum_{ll'} \ket{^{l'}_{}} \prescript{}{\lceil N \rceil^*}{\bra{^{l}}} \otimes \sum_{jj',qq'} {^S \textbf{H}^{jj'}} {^S\textbf{B}_{l'l, j'j}^{\;\;\; \; \;\; \; \;\;qq'}} \ket{^{}_{q}} \prescript{}{ \lceil S \rceil}{\bra{^{}_{q'}}} \\
& = \sum_{ll', jj', qq'} {^S\textbf{H}^{jj'}} {^S\textbf{B}_{l'l, j'j}^{\;\;\; \; \;\; \; \;\;qq'}} \ket{^{l'}_{ \; \; q}} \prescript{}{ \lceil N \rceil^* \otimes \lceil S \rceil}{\bra{^{l}_{ \; \; q'}}}
\end{align}

\begin{align}
& \llbracket z_{0} \triangleleft \lambda^{l} x_{1}.{}^{c^1}t((x_{2} \triangleleft y_{2})\triangleright({}^{\vee}{}^{\cup}x_{1}\triangleright z_{2})) \rrbracket_{g^S} = 
\Tr_{\tilde{S}} \left( \Tr_{\tilde{N}} \left( \llbracket z_{0} \rrbracket_{g^S} . \llbracket\lambda^{l} x_{1}.{}^{c^1}t((x_{2} \triangleleft y_{2})\triangleright({}^{\vee}{}^{\cup}x_{1}\triangleright z_{2})) \rrbracket_{g^S} \right) \right) \nonumber \\
& =  \Tr_{\tilde{S}} \left( \Tr_{\tilde{N}} \left( \sum_{kk', tt',mm', nn'} {^S\textbf{D}_{k'k\;m'm}^{\; \; \; tt'\; \; \;nn'}} \ket{^{k'\;m'}_{\; \; \; t\; \; \;n} } \prescript{}{\lceil N \rceil^* \otimes \lceil N \rceil \otimes (\lceil S \rceil^* \otimes \lceil N \rceil)}{\bra{^{k\;m}_{\; \;t'\; \;n'\;}}}  \right. \right. \nonumber \\
& \left. \left. \cdot \sum_{ll', jj', qq'} {^S\textbf{H}^{jj'}} {^S\textbf{B}_{l'l, j'j}^{\;\;\; \; \;\; \; \;\;qq'}} \ket{^{l'}_{ \; \; q}} \prescript{}{ \lceil N \rceil^* \otimes \lceil S \rceil}{\bra{^{l}_{ \; \; q'}}} \right) \right) \nonumber \\
& = \sum_{kk', tt',mm', nn', jj'} {^S\textbf{D}_{k'k\;m'm}^{\; \; \; tt'\; \; \;nn'}} {^S\textbf{H}^{jj'}} {^S\textbf{B}_{n'n, j'j}^{\;\;\; \; \;\; \; \;\;mm'}}  \ket{^{k'}_{\; \; \; t} } \prescript{}{\lceil N \rceil^* \otimes \lceil N \rceil}{\bra{^{k}_{\; \;t'\;}}}
\end{align}

\begin{align}
&\llbracket y_{0}\triangleright (z_{0} \triangleleft \lambda^{l} x_{1}.{}^{c^1}t((x_{2} \triangleleft y_{2})\triangleright({}^{\vee}{}^{\cup}x_{1}\triangleright z_{2}))) \rrbracket_{g^S} = \Tr_{\tilde{N}} \left(  \llbracket y_{0} \rrbracket_{g^S} \cdot \llbracket z_{0} \triangleleft \lambda^{l} x_{1}.{}^{c^1}t((x_{2} \triangleleft y_{2})\triangleright({}^{\vee}{}^{\cup}x_{1}\triangleright z_{2})) \rrbracket_{g^S}  \right) \nonumber \\
& = \Tr_{\tilde{N}} \left( \sum_{rr'} {^S\textbf{M}}^{rr'} \ket{_{r}}\prescript{}{\lceil N \rceil}{\bra{_{r'}}} \cdot \sum_{kk', tt',mm', nn', jj'} {^S\textbf{D}_{k'k\;m'm}^{\; \; \; tt'\; \; \;nn'}} {^S\textbf{H}^{jj'}} {^S\textbf{B}_{n'n, j'j}^{\;\;\; \; \;\; \; \;\;mm'}}  \ket{^{k'}_{\; \; \; t}} \prescript{}{\lceil N \rceil^* \otimes \lceil N \rceil}{\bra{^{k}_{\; \;t'\;}}} \right) \nonumber\\
& =  \sum_{rr', tt',mm', nn', jj'} {^S\textbf{M}}^{rr'} {^S\textbf{D}_{r'r\;m'm}^{\; \; \; tt'\; \; \;nn'}} {^S\textbf{H}^{jj'}} {^S\textbf{B}_{n'n, j'j}^{\;\;\; \; \;\; \; \;\;mm'}} \ket{^{}_{t}} \prescript{}{ \lceil N \rceil}{\bra{^{}_{t'}}}\\
&= \llbracket \text{man\_die\_de\_hond\_bijt'} \rrbracket^2_{I^S}.
\end{align}

\subsection{Interpretations in $\mathfrak{S}$:}\label{spininter}

\subsubsection{Subject Relativization}

We start by using the interpretations of variables in the interpretation of $E_\backslash$ as given in eq. \ref{triangleright}, which are particular forms of eq. \ref{conv}.  The variables can have any value with the only requirement that it is neither $\rho_0$ nor $\rho_1$. This is because the resulting states must have a non-zero probability of being projected on either of these states, which is necessary for the following step.

\begin{align}
    \llbracket (x_2 \triangleleft y_2) \triangleright z_2 \rrbracket_{g^\mathfrak{S}}& = \llbracket z_2 \rrbracket_{g^\mathfrak{S}} * \llbracket x_2 \triangleleft y_2 \rrbracket_{g^\mathfrak{S}} = \frac{\left(\left\llbracket  x_2 \triangleleft y_2 \right\rrbracket_{g^\mathfrak{S}} \right)^\frac{1}{2} \cdot \left\llbracket z_2 \right\rrbracket_{g^\mathfrak{S}} \cdot \left(\left\llbracket  x_2 \triangleleft y_2 \right\rrbracket_{g^\mathfrak{S}}\right)^\frac{1}{2}}{\Tr_\mathfrak{S} \left( \left(\left\llbracket  x_2 \triangleleft y_2 \right\rrbracket_{g^\mathfrak{S}} \right)^\frac{1}{2} \cdot \left\llbracket z_2 \right\rrbracket_{g^\mathfrak{S}} \cdot \left(\left\llbracket  x_2 \triangleleft y_2 \right\rrbracket_{g^\mathfrak{S}}\right)^\frac{1}{2} \right)}.
\end{align}

\begin{align}\label{mod1}
 & \llbracket x \triangleright ((x_2 \triangleleft y_2) \triangleright z_2) \rrbracket_{g^\mathfrak{S}} =   \left( \llbracket z_2 \rrbracket_{g^\mathfrak{S}} * \llbracket x_2 \triangleleft y_2 \rrbracket_{g^\mathfrak{S}}\right) * \llbracket x \rrbracket_{g^\mathfrak{S}}  \nonumber  \\
 &=\frac{\left( \left\llbracket  x \right\rrbracket_{g^\mathfrak{S}} \right)^\frac{1}{2} \cdot \left(\llbracket z_2 \rrbracket_{g^\mathfrak{S}} * \llbracket x_2 \triangleleft y_2 \rrbracket_{g^\mathfrak{S}}\right)   \cdot \left(\left\llbracket  x \right\rrbracket_{g^\mathfrak{S}}\right)^\frac{1}{2}}{\Tr_\mathfrak{S} \left( \left( \left\llbracket  x \right\rrbracket_{g^\mathfrak{S}} \right)^\frac{1}{2} \cdot \left(\llbracket z_2 \rrbracket_{g^\mathfrak{S}} * \llbracket x_2 \triangleleft y_2 \rrbracket_{g^\mathfrak{S}}\right)   \cdot \left(\left\llbracket  x \right\rrbracket_{g^\mathfrak{S}}\right)^\frac{1}{2} \right)} 
\end{align}

Looking at the interpretation of \textit{xleft} in eq. \ref{xleftprev}, we first work out eq. \ref{projuni} with $\llbracket x \rrbracket_{g^\mathfrak{S}}$ substituted by $\llbracket ^{\vee \cup} x_1 \rrbracket =  T'^0_\square \left( T'^0_\lozenge \left( \llbracket x_1 \rrbracket_{g^\mathfrak{S}} \right) \right)$ because of assignment $g^\mathfrak{S}_{x,^{\vee \cup} x_1}$,  and with $\llbracket x_1 \rrbracket_{g^\mathfrak{S}}$ substituted by $I$ in its turn, because of the assignment $g^{\mathfrak{S}}_{x,I}$. Recall that in our definitions $U_0=\mathbb{1}$. Since controlled commutativity is not used, $n=0$ and $\left(S_+\right)^0=\mathbb{1}$. In both steps below,  pure state $\rho_0$ will be preserved, taking into account that
\begin{equation} \label{convpar}
    \llbracket t^A \rrbracket_{g^\mathfrak{S}} * \llbracket u^B \rrbracket_{g^\mathfrak{S}} =  \llbracket u^B \rrbracket_{g^\mathfrak{S}},
\end{equation} when $\llbracket u^B \rrbracket_{g^\mathfrak{S}}$ equals $\rho_0$ or $\rho_1$. To show this, take $\llbracket t^A \rrbracket_{g^\mathfrak{S}} = \begin{pmatrix}
a & b\\ 
c & d
\end{pmatrix}$ and $\llbracket u^B \rrbracket_{g^\mathfrak{S}}=\ket{0} \prescript{}{\mathfrak{S}}{\bra{0}} = \begin{pmatrix}
0 & 0\\ 
0 & 1
\end{pmatrix}$, 

\begin{equation}\label{convpure}
\llbracket t^A \rrbracket_{g^\mathfrak{S}} * \llbracket u^B \rrbracket_{g^\mathfrak{S}} = \frac{\begin{pmatrix}
0 & 0\\ 
0 & 1
\end{pmatrix} \begin{pmatrix}
a & b\\ 
c & d
\end{pmatrix} \begin{pmatrix}
0 & 0\\ 
0 & 1
\end{pmatrix}}{\Tr \left( \begin{pmatrix}
0 & 0\\ 
0 & 1
\end{pmatrix} \begin{pmatrix}
a & b\\ 
c & d
\end{pmatrix} \begin{pmatrix}
0 & 0\\ 
0 & 1
\end{pmatrix} \right)} = \frac{\begin{pmatrix}
0 & 0\\ 
0 & d
\end{pmatrix}}{d} = \begin{pmatrix}
0 & 0\\ 
0 & 1
\end{pmatrix},
\end{equation} and similarly for $\llbracket u^B \rrbracket_{g^\mathfrak{S}}=\rho_1$.

Therefore, the concrete interpretation of the \textit{xleft} rule uses

\begin{align}
   \llbracket ^{\vee \cup} x_1 \rrbracket =  T'^0_\square \left( T'^0_\lozenge \left( I \right) \right) = I * \ket{0} \prescript{}{\mathfrak{S}}{\bra{0}} = \ket{0} \prescript{}{\mathfrak{S}}{\bra{0}},
\end{align} which substituted in eq.\ref{mod1} gives

\begin{align}
&\llbracket \lambda^{l} x_{1}.{}^{c^0}t({}^{\vee}  {}^{\cup} x_{1}\triangleright ((x_{2} \triangleleft  y_{2})\triangleright z_{2})) \rrbracket_{g^\mathfrak{S}} = \nonumber  \\
&= S^0_+ \left(\left( \llbracket z_2 \rrbracket_{g^\mathfrak{S}} * \llbracket x_2 \triangleleft y_2 \rrbracket_{g^\mathfrak{S}}  \right) * \ket{0}\prescript{}{\mathfrak{S}}{\bra{0}} \right) \left(S^0_+ \right)^\dagger  \nonumber \\
& =\ket{0}\prescript{}{\mathfrak{S}}{\bra{0}}
\end{align}

In the following two steps, the interpretations of rules $E_\slash$ and $E_\backslash$ are used. In the last step of \ref{something}, we refer again to eq.\ref{subrel}.

\begin{align} \label{something}
& \llbracket z_{0} \triangleleft \lambda^{l} x_{1}.{}^{c^0}t({}^{\vee}  {}^{\cup} x_{1}\triangleright ((x_{2} \triangleleft  y_{2})\triangleright z_{2})) \rrbracket_{g^\mathfrak{S}}  =  \llbracket z_{0} \rrbracket_{g^\mathfrak{S}} * \llbracket \lambda^{l} x_{1}.{}^{c^0}t({}^{\vee}  {}^{\cup} x_{1}\triangleright ((x_{2} \triangleleft  y_{2})\triangleright z_{2})) \rrbracket_{g^\mathfrak{S}} \nonumber \\
& =  \llbracket z_{0} \rrbracket_{g^\mathfrak{S}} * \ket{0}\prescript{}{\mathfrak{S}}{\bra{0}} = \frac{\left( \ket{0}\prescript{}{\mathfrak{S}}{\bra{0}} \right )^\frac{1}{2} \cdot \left\llbracket  z_0 \right\rrbracket_{g^\mathfrak{S}} \cdot \left(\ket{0}\prescript{}{\mathfrak{S}}{\bra{0}}\right)^\frac{1}{2}}{\Tr_\mathfrak{S} \left(\left( \ket{0}\prescript{}{\mathfrak{S}}{\bra{0}} \right )^\frac{1}{2} \cdot \left\llbracket  z_0 \right\rrbracket_{g^\mathfrak{S}} \cdot \left(\ket{0}\prescript{}{\mathfrak{S}}{\bra{0}}\right)^\frac{1}{2} \right)} = \ket{0}\prescript{}{\mathfrak{S}}{\bra{0}} \nonumber \\
&= \llbracket \text{die\_de\_hond\_bijt'} \rrbracket^1_{I^\mathfrak{S}}
\end{align}

\begin{align}
&\llbracket y_{0}\triangleright (z_{0} \triangleleft \lambda^{l} x_{1}.{}^{c^0}t({}^{\vee}  {}^{\cup} x_{1}\triangleright ((x_{2} \triangleleft  y_{2})\triangleright z_{2}))) \rrbracket_{g^\mathfrak{S}} =  \llbracket (z_{0} \triangleleft \lambda^{l} x_{1}.{}^{c^0}t({}^{\vee}  {}^{\cup} x_{1}\triangleright ((x_{2} \triangleleft  y_{2})\triangleright z_{2}))) \rrbracket_{g^\mathfrak{S}} * \llbracket y_{0} \rrbracket_{g^\mathfrak{S}}  \nonumber \\
& = \ket{0}\prescript{}{\mathfrak{S}}{\bra{0}} *   \left\llbracket  y_0 \right\rrbracket_{g^\mathfrak{S}} = \frac{\left( \left\llbracket  y_0 \right\rrbracket_{g^\mathfrak{S}} \right )^\frac{1}{2} \cdot \ket{0}\prescript{}{\mathfrak{S}}{\bra{0}}  \cdot \left(\left\llbracket  y_0 \right\rrbracket_{g^\mathfrak{S}}\right)^\frac{1}{2}}{\Tr_\mathfrak{S} \left(\left( \left\llbracket  y_0 \right\rrbracket_{g^\mathfrak{S}} \right )^\frac{1}{2} \cdot \ket{0}\prescript{}{\mathfrak{S}}{\bra{0}}  \cdot \left(\left\llbracket  y_0 \right\rrbracket_{g^\mathfrak{S}}\right)^\frac{1}{2} \right)} \nonumber \\
&= \llbracket \text{man\_die\_de\_hond\_bijt'} \rrbracket^1_{I^\mathfrak{S}}.
\end{align}

\subsubsection{Object Relativization}

Just as in the previous derivations, once more we use the interpretations of $E_\backslash$ in the two first steps.

\begin{align}
 \llbracket x \triangleright z_2 \rrbracket_{g^\mathfrak{S}}& = \llbracket z_2 \rrbracket_{g^\mathfrak{S}} *  \llbracket x \rrbracket_{g^\mathfrak{S}} = \frac{\left(  \left\llbracket  x \right\rrbracket_{g^\mathfrak{S}} \right)^\frac{1}{2} \cdot \left\llbracket z_2 \right\rrbracket_{g^\mathfrak{S}} \cdot \left( \left\llbracket  x \right\rrbracket_{g^\mathfrak{S}}\right)^\frac{1}{2}}{\Tr_\mathfrak{S} \left( \left(  \left\llbracket  x \right\rrbracket_{g^\mathfrak{S}} \right)^\frac{1}{2} \cdot \left\llbracket z_2 \right\rrbracket_{g^\mathfrak{S}} \cdot \left( \left\llbracket  x \right\rrbracket_{g^\mathfrak{S}}\right)^\frac{1}{2} \right)}.
\end{align}

 \begin{align}
& \llbracket (x_{2} \triangleleft y_{2})\triangleright(x \triangleright z_{2}) \rrbracket_{g^\mathfrak{S}} =  \left( \llbracket z_2 \rrbracket_{g^\mathfrak{S}} *  \llbracket x \rrbracket_{g^\mathfrak{S}}  \right) * \llbracket x_2 \triangleleft y_2 \rrbracket_{g^\mathfrak{S}} \nonumber \\
 & = \frac{\left( \llbracket x_2 \triangleleft y_2 \rrbracket_{g^\mathfrak{S}} \right)^\frac{1}{2} \cdot \left( \llbracket z_2 \rrbracket_{g^\mathfrak{S}} *  \llbracket x \rrbracket_{g^\mathfrak{S}}  \right) \cdot \left( \llbracket x_2 \triangleleft y_2 \rrbracket_{g^\mathfrak{S}} \right)^\frac{1}{2}}{\Tr_\mathfrak{S} \left(\left( \llbracket x_2 \triangleleft y_2 \rrbracket_{g^\mathfrak{S}} \right)^\frac{1}{2} \cdot \left( \llbracket z_2 \rrbracket_{g^\mathfrak{S}} *  \llbracket x \rrbracket_{g^\mathfrak{S}}  \right) \cdot \left( \llbracket x_2 \triangleleft y_2 \rrbracket_{g^\mathfrak{S}} \right)^\frac{1}{2} \right)} 
 \end{align}

In the application of the interpretation of $xleft$ in eq.\ref{xleftfirst}) is the same as in the previous reading, except that controlled commutation is used once, so that $m=1$, meaning that $\left(S_+\right)^1=S_+$, $U_0=\mathbb{1}$ :

\begin{align}
   \llbracket ^{\vee \cup} x_1 \rrbracket =  T'^0_\square \left( T'^0_\lozenge \left( I \right) \right) = I * \ket{0} \prescript{}{\mathfrak{S}}{\bra{0}} = \ket{0} \prescript{}{\mathfrak{S}}{\bra{0}},
\end{align} which substituted in \ref{mod1} gives

\begin{align}
&\llbracket \lambda^{l} x_{1}.{}^{c^1}t (x_{2} \triangleleft y_{2})\triangleright( ^{\vee \cup} x_1 \triangleright z_{2})) \rrbracket_{g^\mathfrak{S}} = \nonumber  \\
&= \frac{S_+ \left( \left( \llbracket z_2 \rrbracket_{g^\mathfrak{S}} *  \ket{0} \prescript{}{\mathfrak{S}}{\bra{0}} \right) * \llbracket x_2 \triangleleft y_2 \rrbracket_{g^\mathfrak{S}} \right) \left(S_+ \right)^\dagger}{\Tr_\mathfrak{S} \left( S_+ \left( \left( \llbracket z_2 \rrbracket_{g^\mathfrak{S}} *  \ket{0} \prescript{}{\mathfrak{S}}{\bra{0}} \right) * \llbracket x_2 \triangleleft y_2 \rrbracket_{g^\mathfrak{S}} \right) \left(S_+ \right)^\dagger \right)}  \nonumber \\
& =\ket{1}\prescript{}{\mathfrak{S}}{\bra{1}},
\end{align} since \begin{equation}\label{convpure}
\frac{S_+ \llbracket t^A \rrbracket_g^{\mathfrak{S}} S_+^\dagger}{\Tr_\mathfrak{S} \left(S_+ \llbracket t^A \rrbracket_g^{\mathfrak{S}} S_+^\dagger \right)}=
\frac{\begin{pmatrix}
0 & 1\\ 
0 & 0
\end{pmatrix} \begin{pmatrix}
a & b\\ 
c & d
\end{pmatrix} \begin{pmatrix}
0 & 0\\ 
1 & 0
\end{pmatrix}}{\Tr \left( \begin{pmatrix}
0 & 1\\ 
0 & 0
\end{pmatrix} \begin{pmatrix}
a & b\\ 
c & d
\end{pmatrix} \begin{pmatrix}
0 & 0\\ 
1 & 0
\end{pmatrix} \right)} = \frac{\begin{pmatrix}
d & 0\\ 
0 & 0
\end{pmatrix}}{d} = \begin{pmatrix}
1 & 0\\ 
0 & 0
\end{pmatrix} = \ket{1} \prescript{}{\mathfrak{S}}{\bra{1}}.
\end{equation}

Finally, for the interpretations of $E_\slash$ and $E_\backslash$:

\begin{align}
& \llbracket z_{0} \triangleleft \lambda^{l} x_{1}.{}^{c^1}t((x_{2} \triangleleft y_{2})\triangleright({}^{\vee}{}^{\cup}x_{1}\triangleright z_{2})) \rrbracket_{g^\mathfrak{S}}  =  \llbracket z_{0} \rrbracket_{g^\mathfrak{S}} * \llbracket \lambda^{l} x_{1}.{}^{c^1}t((x_{2} \triangleleft y_{2})\triangleright({}^{\vee}{}^{\cup}x_{1}\triangleright z_{2})) \rrbracket_{g^\mathfrak{S}} \nonumber \\
& = \frac{\left( \ket{1}\prescript{}{\mathfrak{S}}{\bra{1}} \right )^\frac{1}{2} \cdot \left\llbracket  z_0 \right\rrbracket_{g^\mathfrak{S}} \cdot \left(\ket{1}\prescript{}{\mathfrak{S}}{\bra{1}}\right)^\frac{1}{2}}{\Tr_\mathfrak{S} \left(\left( \ket{1}\prescript{}{\mathfrak{S}}{\bra{1}} \right )^\frac{1}{2} \cdot \left\llbracket  z_0 \right\rrbracket_{g^\mathfrak{S}} \cdot \left(\ket{1}\prescript{}{\mathfrak{S}}{\bra{1}}\right)^\frac{1}{2} \right)} = \ket{1}\prescript{}{\mathfrak{S}}{\bra{1}} \nonumber \\
&= \llbracket \text{die\_de\_hond\_bijt'} \rrbracket^2_{I^\mathfrak{S}}.
\end{align}

\begin{align}
&\llbracket y_{0}\triangleright (z_{0} \triangleleft \lambda^{l} x_{1}.{}^{c^1}t((x_{2} \triangleleft y_{2})\triangleright({}^{\vee}{}^{\cup}x_{1}\triangleright z_{2}))) \rrbracket_{g^\mathfrak{S}} = \llbracket z_{0} \triangleleft \lambda^{l} x_{1}.{}^{c^1}t((x_{2} \triangleleft y_{2})\triangleright({}^{\vee}{}^{\cup}x_{1}\triangleright z_{2}))  \rrbracket_{g^\mathfrak{S}} *  \llbracket y_{0} \rrbracket_{g^\mathfrak{S}} \nonumber \\
& = \frac{\left( \left\llbracket  y_0 \right\rrbracket_{g^\mathfrak{S}} \right )^\frac{1}{2} \cdot \ket{1}\prescript{}{\mathfrak{S}}{\bra{1}}  \cdot \left(\left\llbracket  y_0 \right\rrbracket_{g^\mathfrak{S}}\right)^\frac{1}{2}}{\Tr_\mathfrak{S} \left(\left( \left\llbracket  y_0 \right\rrbracket_{g^\mathfrak{S}} \right )^\frac{1}{2} \cdot \ket{1}\prescript{}{\mathfrak{S}}{\bra{1}}  \cdot \left(\left\llbracket  y_0 \right\rrbracket_{g^\mathfrak{S}}\right)^\frac{1}{2}\right)} \nonumber \\
&= \llbracket \text{man\_die\_de\_hond\_bijt'} \rrbracket^2_{I^\mathfrak{S}}.
\end{align}

\section{Proof transformation: beta reduction} \label{betareduction}

The $\beta$-reduction is one of the rewrite rules of the $\lambda$-calculus. It asserts that applying a term with a lambda-bound variable to a certain argument is equivalent to substituting that argument directly in the original term, before introducing the lambda. In proof-theoretic terms, if an introduction rule is used followed by an elimination rule, the derivation is not minimal. To elucidate this point, below is the skeleton of a derivation where a term of type $A$ is proved twice, by axiom and by an unknown proof:

$$\prftree[straight][r]{$\backslash _E$}{\prftree[straight][r]{}{\vdots}{\Delta \vdash n:A}}{\prftree[straight][r]{$\backslash _I$}{\prftree[straight][r]{}{\prftree[straight][r]{}{\prftree[straight][r]{$_{axiom}$}{}{x:A \vdash x:A}}{\vdots}}{x:A,\Gamma \vdash t:B}}{\Gamma \vdash\lambda^l x. m: A\backslash B }}{(\Gamma, \Delta)\vdash n \triangleright (\lambda^l x.m)  :B }.$$
\\ The $\beta$ reduction consists of substituting the unknown proof of the term of type $A$ in place of the axiom, reducing the need for the double proof of that term, and consequently the size of the proof:

$$\prftree[straight][r]{}{\prftree[straight][r]{}{\prftree[straight][r]{}{\vdots}{\Delta \vdash n:A}}{\vdots}}{\Delta, \Gamma \vdash m[x/n]:B}.$$ Through this reduction, a map from one conclusion to the other can be obtained, which has to be an equality regarding their interpretations:

$$\llbracket n \triangleright (\lambda^l x.m)] \rrbracket_g = \llbracket m[x/n] \rrbracket_g, \forall g.$$ This equality will be used to check that the density matrix construction interpretation is consistent with the $\lambda$-calculus. Below a concrete symbolic derivation before the reduction is shown:

$$\scalebox{.85}{\prftree[straight][r]{$\backslash _{E_3}$}{\prftree[straight][r]{$\backslash _{E_2}$}{\prftree[straight][r]{$_{ax}$}{}{w:B \vdash w:B}}{\prftree[straight][r]{$_{ax}$}{}{z: B\backslash (A/B) \vdash z: B\backslash (A/B)}}{ w:B, z:B\backslash (A/B) \vdash (w \triangleright z) : A/B}}{\prftree[straight][r]{$\backslash _{I_1}$}{\prftree[straight][r]{$/ _{E_1}$}{\prftree[straight][r]{$_{ax}$}{}{x:A/B \vdash x: A/B}}{\prftree[straight][r]{$_{ax}$}{}{y:B \vdash y:B}}{x:A/B, y:B \vdash (x \triangleleft y):A}}{ y:B \vdash \lambda^l x. (x \triangleleft y): (A/B)\backslash A}}{(w:B, z:B\backslash (A/B), u:B) \vdash (w \triangleright z) \triangleright (\lambda^l x. (x \triangleleft y)) :A}.}$$

The interpretation in the spatial space $S$ of the several steps of the proof is given below, following the numbering in the proof:

$$E_{/_1}: \; \llbracket (x \triangleleft y) \rrbracket_{g^S} = \sum_{ii', jj'} \prescript{}{}{{^S\textbf{X}}_{\; \; jj'}^{i'i}} \prescript{}{}{ {^S\textbf{Y}}^{j'j}}  \ket{^{}_{i}} \prescript{}{ \lceil A \rceil}{\bra{^{}_{i'}}},$$

$$I_{\backslash_1}: \; \llbracket  \lambda^l x. (x \triangleleft y)\rrbracket_{g^S} = \sum_{ii', jj'} \ket{_{j}^{\;i'}}\prescript{}{\lceil B \rceil \otimes \lceil A \rceil^*}{\bra{_{j'}^{\;i}}} \otimes\prescript{}{}{{^S\textbf{Y}}^{j'j}}  \ket{^{}_{i}} \prescript{}{ \lceil A \rceil}{\bra{^{}_{i'}}},$$

$$E_{\backslash _2}: \; \llbracket (w \triangleright z) \rrbracket_g = \sum_{ll',mm', nn'} \prescript{}{}{{^S\textbf{W}}^{ll'}} \prescript{}{}{{^S\textbf{Z}}_{l'l, nn'}^{ \; \;  m'm}}   \ket{_{m}^{\; \;  n'}} \prescript{}{ \lceil A \rceil \otimes \lceil B \rceil^* }{\bra{_{m'}^{\; \;  n}}},$$

$$E_{\backslash _3}: \; \llbracket  (w \triangleright z) \triangleright (\lambda^l x. (x \triangleleft y)) \rrbracket_{g^S} = \sum_{ii', jj', ll'}   \prescript{}{}{{^S\textbf{W}}^{ll'}} \prescript{}{}{{^S\textbf{Z}}_{l'l, jj'}^{ \; \; i'i }} \prescript{}{}{{^S\textbf{Y}}^{j'j}}  \ket{^{}_{i}} \prescript{}{ \lceil A \rceil}{\bra{^{}_{i'}}}.$$

In spin space $\mathfrak{S}$ the interpretation of the proof steps is as follows:

$$E_{/_1}: \; \llbracket (x \triangleleft y) \rrbracket_{g^\mathfrak{S}} = \llbracket x \rrbracket_{g^\mathfrak{S}} * \llbracket y \rrbracket_{g^\mathfrak{S}} $$

$$I_{\backslash_1}: \; \llbracket  \lambda^l x. (x \triangleleft y)\rrbracket_{g^\mathfrak{S}} = I * \llbracket y \rrbracket_{g^\mathfrak{S}} = \llbracket y \rrbracket_{g^\mathfrak{S}}$$

$$E_{\backslash _2}: \; \llbracket (w \triangleright z) \rrbracket_g = \llbracket z \rrbracket_{g^\mathfrak{S}} * \llbracket w \rrbracket_{g^\mathfrak{S}}$$

$$E_{\backslash _3}: \; \llbracket  (w \triangleright z) \triangleright (\lambda^l x. (x \triangleleft y)) \rrbracket_{g^S} =  \llbracket y \rrbracket_{g^\mathfrak{S}} * \left( \llbracket z \rrbracket_{g^\mathfrak{S}} * \llbracket w \rrbracket_{g^\mathfrak{S}} \right)   $$

A similar treatment is done for the derivation after the reduction:

$$\prftree[straight][r]{$\backslash _{E_4}$}{\prftree[straight][r]{$\backslash _{E_2}$}{\prftree[straight][r]{$_{ax}$}{}{w:B \vdash w:B}}{\prftree[straight][r]{$_{ax}$}{}{z: B\backslash (A/B) \vdash z: B\backslash (A/B)}}{ w:B, z:B\backslash (A/B) \vdash (w \triangleright z) : A/B}}{\prftree[straight][r]{$_{ax}$}{}{y:B \vdash y:B}}{w:B, z:B\backslash (A/B), u:B\vdash ((w \triangleright z) \triangleleft y) :A}.$$ The value of $\llbracket (w \triangleright z) \rrbracket_{g}$ is the same as before. For $\llbracket ((w \triangleright z) \triangleleft y) \rrbracket_{g^S}$:

$$E_{\backslash _4}: \; \llbracket  ((w \triangleright z) \triangleleft y) \rrbracket_{g^S} = \sum_{ii', jj', ll'}   {^S\textbf{W}}^{ll'} {^S\textbf{Z}}_{l'l, jj'}^{ \; \; i'i } {^S\textbf{Y}}^{j'j}  \ket{^{}_{i}} \prescript{}{ \lceil A \rceil}{\bra{^{}_{i'}}}.$$

On the spin space, we have

$$E_{\backslash _4}: \; \llbracket  ((w \triangleright z) \triangleleft y) \rrbracket_{g^\mathfrak{S}} =\llbracket y \rrbracket_{g^\mathfrak{S}} *  \left(  \llbracket z \rrbracket_{g^\mathfrak{S}} * \llbracket w \rrbracket_{g^\mathfrak{S}} \right)$$

Comparing the two derivations and interpretations, the conclusion is that $$\llbracket E_{\backslash _4} (y, z(w)) \rrbracket_{g^S} = \llbracket E_{\backslash _3} (z(w), \lambda x. x(y)) \rrbracket_{g^S},$$ as expected, and 

$$\llbracket E_{\backslash _4} (y, z(w)) \rrbracket_{g^\mathfrak{S}} = \llbracket E_{\backslash _3} (z(w), \lambda x. x(y)) \rrbracket_{g^\mathfrak{S}}.$$ 

\end{document}